%% file: main.tex
\documentclass{article}
\usepackage[T1]{fontenc}
\usepackage{iclr2027_conference,times}
\usepackage[zerostyle=a]{newtxtt}
\usepackage{amsmath,amssymb}
\usepackage{booktabs,multirow,array,tabularx}
\usepackage{graphicx}
\usepackage{tikz}
\usetikzlibrary{arrows.meta,calc}
\usepackage{enumitem}
\usepackage{placeins}
\usepackage{caption}
\usepackage{hyperref}
\usepackage{xurl}
\usepackage[capitalize,noabbrev]{cleveref}
\graphicspath{{images/}}
\DeclareMathSizes{5.9}{5.9}{5}{5}
\DeclareMathSizes{6.8}{6.8}{5}{5}

\newif\ifanonymous
\anonymousfalse
\ifanonymous\else\iclrfinalcopy\fi
\hypersetup{pdftitle={The Geometry of Lies: Knowledge and Persona Directions for Detecting Deception in
  Large Language Models}}
\ifanonymous\else
\hypersetup{pdfauthor={Md. Abdullah Al Zami, Suraiya Shanaz, Mahdi Hasan, Shoaib Ahmed Dipu, Farig Yousuf
  Sadeque, Md. Sabbir Ahmed}}
\fi

\makeatletter
\newcommand{\setV}[2]{\expandafter\def\csname num@#1\endcsname{#2}}
\newcommand{\V}[1]{\ifcsname num@#1\endcsname\csname num@#1\endcsname\else
  \PackageError{numbers}{Undefined number `#1'}{The key is missing from the generated numbers files}\fi}
\makeatother
\input{generated/numbers.tex}
\input{generated/numbers_xmodel.tex}
\input{generated/numbers_arxiv.tex}

\newsavebox{\fitbox}
\newcommand{\fitwidth}[1]{\sbox{\fitbox}{#1\unskip}%
  \ifdim\wd\fitbox>\linewidth\resizebox{\linewidth}{!}{\usebox{\fitbox}}\else\usebox{\fitbox}\fi}
\newcommand{\twoline}[1]{\begin{tabular}[t]{@{}c@{}}#1\end{tabular}}
\newcommand{\twolinel}[1]{\begin{tabular}[t]{@{}l@{}}#1\end{tabular}}
\newcommand{\stackc}[1]{\begin{tabular}[b]{@{}c@{}}#1\end{tabular}}
\newcommand{\stackl}[1]{\begin{tabular}[b]{@{}l@{}}#1\end{tabular}}

\newcommand{\hpkr}{\mbox{HP-KR}}
\newcommand{\hpc}{\mbox{HP-C}}
\title{The Geometry of Lies: Knowledge and\\ Persona Directions for Detecting Deception\\ in Large Language
Models}

\author{\mdseries Md.\ Abdullah Al Zami, Suraiya Shanaz, Mahdi Hasan, \\
Shoaib Ahmed Dipu, Farig Yousuf Sadeque \& Md.\ Sabbir Ahmed \\
Department of Computer Science and Engineering \\
BRAC University \\
Dhaka, Bangladesh \\
\texttt{abdullah.al.zami@g.bracu.ac.bd}, \texttt{suraiya.shanaz@g.bracu.ac.bd}, \\
\texttt{mahdi.hasan.meem@g.bracu.ac.bd}}

\begin{document}

\maketitle

\begin{abstract}
\input{sections/abstract}
\end{abstract}

\input{sections/introduction}
\input{sections/related}
\input{sections/method}
\input{sections/results_llama}
\input{sections/results_models}
\input{sections/results_external}
\input{sections/discussion}

\input{sections/statements}

\bibliography{refs}
\bibliographystyle{iclr2027_conference}

\appendix
\input{sections/appendix}

\end{document}

%% file: generated/numbers.tex
\setV{ndec-hpkr}{139}
\setV{bank-new}{10}
\setV{zs-dec-hpkr-L20}{0.148}
\setV{zs-dec-hpkr-L40}{0.593}
\setV{zs-ext-lb-L40}{0.899}
\setV{off-ext-L20}{0.904}
\setV{off-base-L40}{0.732}
\setV{diag-base-L40}{0.983}
\setV{off-dec-L40}{0.832}
\setV{off-ext-L40}{0.916}
\setV{offx-ext-L40}{0.917}
\setV{off-pcthree-L40}{0.639}
\setV{op-den-balacc-hpkr-L40}{0.910}
\setV{bp-auroc-hpkr-L20}{0.109}
\setV{bp-auroc-hpkr-L40}{0.936}
\setV{bp-auroc-lb-L40}{0.778}
\setV{lb-mean-auroc-hpkr}{0.12}
\setV{lb-mean-auroc-avg}{0.682}
\setV{lb-examples}{72,863}
\setV{depth-n}{16}
\setV{depth-off-ext-peak}{0.927}
\setV{depth-off-ext-peak-layer}{36}
\setV{depth-hpkr-den-balacc-peak}{0.910}
\setV{depth-hpkr-den-balacc-peak-layer}{40}
\setV{depth-hpkr-den-balacc-L36}{0.745}
\setV{depth-hpkr-den-balacc-L44}{0.757}
\setV{depth-hpkr-base-auroc-peak}{0.967}
\setV{depth-hpkr-base-auroc-peak-layer}{36}
\setV{th-aisi-ext-L20}{0.105}
\setV{th-aisi-gain-L40}{0.100}
\setV{th-aisi-gain-lo-L40}{0.086}
\setV{th-aisi-gain-hi-L40}{0.113}
\setV{th-aisi-ext-L40}{0.021}
\setV{th-aisi-ext-lo-L40}{0.006}
\setV{th-aisi-ext-hi-L40}{0.035}
\setV{th-aisi-train-rows}{3,285}
\setV{th-aisi-train-lies}{1,634}
\setV{th-aisi-val-lies}{1,730}
\setV{th-aisi-val-honest}{498}
\setV{th-uni-off-mean}{0.657}
\setV{th-dec-off-mean}{0.686}
\setV{th-ext-off-mean}{0.885}
\setV{th-gs-den-L40}{0.436}
\setV{th-gs-ft-best}{0.641}
\setV{th-gs-ft-best-layer}{44}
\setV{th-steer-n}{27}
\setV{th-steer-base}{0.48}
\setV{th-steer-dec-plus-change}{\ensuremath{-}0.28}
\setV{th-steer-ext-plus}{0.20}
\setV{th-steer-ext-plus-change}{\ensuremath{-}0.28}
\setV{th-steer-ext-plus-lo}{\ensuremath{-}0.41}
\setV{th-steer-ext-plus-hi}{\ensuremath{-}0.13}
\setV{th-steer-base-plus-change}{\ensuremath{-}0.09}
\setV{th-steer-base-plus-lo}{\ensuremath{-}0.22}
\setV{th-steer-base-plus-hi}{\ensuremath{+}0.04}
\setV{th-k-known}{60}
\setV{th-k-unknown}{60}
\setV{th-k-rows}{240}
\setV{th-k-verify-orig-correct}{70}
\setV{th-k-style-orig}{95}
\setV{th-k-style-matched}{86}
\setV{th-honest-hpkr}{2,000}
\setV{th-k-ident-lies}{78}
\setV{th-k-ident-honest}{112}
\setV{th-k-den-honest}{135}
\setV{th-k-k-auroc-hpc-L40}{0.830}
\setV{th-k-k-ident-L40}{0.793}
\setV{th-k-k-mean-L40}{0.825}
\setV{th-k-dfive-ident-L40}{0.715}
\setV{th-k-deight-ident-L40}{0.712}
\setV{th-k-deight-flag-lies-L40}{88}
\setV{th-k-deight-flag-hden-L40}{95}
\setV{th-k-deight-flag-hclaims-L40}{0}
\setV{th-k-dnine-auroc-hpc-L40}{0.712}
\setV{th-k-dnine-ident-L40}{0.721}
\setV{th-k-dnine-mean-L40}{0.899}
\setV{th-k-ninek-mean-L40}{0.906}
\setV{th-k-ident-band-L40}{0.733}
\setV{th-k-ident-nowording-L40}{0.792}
\setV{th-k-full-L40}{0.865}
\setV{th-k-full-nowording-L40}{0.349}
\setV{th-k-ident-lo-L40}{0.728}
\setV{th-k-ident-hi-L40}{0.852}
\setV{th-k-probe-ident-L40}{0.779}
\setV{th-rand-hpkr-L40}{0.926}
\setV{th-k-depth-ident-above}{15}
\setV{th-k-depth-n}{16}
\setV{th-k-depth-persona-ident-max}{0.724}
\setV{th-k-gs-ft-kft-L40}{0.861}
\setV{th-k-gs-ft-kft-lo-L40}{0.817}
\setV{th-k-gs-ft-kft-hi-L40}{0.896}
\setV{th-k-gs-ft-ninek-L40}{0.802}
\setV{th-k-gs-ft-ninekft-L40}{0.841}
\setV{th-k-gs-ft-band-L40}{0.824}
\setV{th-k-gs-base-kft-L40}{0.629}
\setV{th-k-gs-gain-L40}{0.294}
\setV{th-k-st-greeting-ft-kft-L40}{0.943}
\setV{th-k-st-greeting-ft-band-L40}{0.922}
\setV{th-k-st-greeting-base-kft-L40}{0.937}
\setV{th-k-st-greeting-base-band-L40}{0.911}
\setV{th-k-st-time-ft-kft-L40}{0.940}
\setV{th-k-st-time-ft-band-L40}{0.870}
\setV{th-k-st-time-base-kft-L40}{0.937}
\setV{th-k-st-time-base-band-L40}{0.918}
\setV{th-k-boot-ident-width}{0.002}
\setV{th-k-rep-criteria}{11}
\setV{th-k-rep-cos-L40}{0.99}
\setV{th-k-op-k-1-L40}{0.21}
\setV{th-k-op-dnine-1-L40}{0.47}
\setV{th-k-op-ninek-1-L40}{0.38}
\setV{th-k-op-probe-1-L40}{0.32}
\setV{th-hk-lie-knows-denies}{125}
\setV{th-hk-denial-auroc}{0.916}
\setV{th-pool-auroc}{0.850}
\setV{th-pool-offx}{0.915}
\setV{th-des-d6-offx}{0.936}
\setV{th-des-d9-offx}{0.917}
\setV{th-pair-ninek-dnine-ident-L40}{0.066}
\setV{th-pair-ninek-dnine-ident-lo-L40}{0.037}
\setV{th-pair-ninek-dnine-ident-hi-L40}{0.098}
\setV{th-pair-ninek-dnine-mean-L40}{0.007}
\setV{th-pair-ninek-dnine-mean-lo-L40}{\ensuremath{-}0.004}
\setV{th-pair-ninek-dnine-mean-hi-L40}{0.018}
\setV{th-k-ninek-off-L20}{0.811}
\setV{th-k-ninek-off-L40}{0.922}
\setV{th-k-k-off-L40}{0.823}
\setV{th-k-ninek-depth-off-mean}{0.853}
\setV{th-k-ninek-depth-lead-min}{0.009}

%% file: generated/numbers_xmodel.tex
\setV{xm-llama-depth-n}{16}
\setV{xm-llama-ci-off-dnine-raw}{0.184}
\setV{xm-llama-ci-off-dnine-raw-lo}{0.172}
\setV{xm-llama-ci-off-dnine-raw-hi}{0.195}
\setV{xm-llama-ci-off-dnine-dfive}{0.084}
\setV{xm-llama-ci-off-ninek-dnine}{0.006}
\setV{xm-llama-ci-off-ninek-dnine-lo}{\ensuremath{-}0.003}
\setV{xm-llama-ci-off-ninek-dnine-hi}{0.015}
\setV{xm-llama-ci-lb-dnine-dfive}{0.101}
\setV{xm-llama-ci-lb-ninek-probe}{0.128}
\setV{xm-llama-roleplay}{69}
\setV{xm-qwen-off-raw}{0.559}
\setV{xm-qwen-off-deight}{0.679}
\setV{xm-qwen-off-dnine}{0.453}
\setV{xm-qwen-off-ninek}{0.514}
\setV{xm-qwen-lb-dnine}{0.441}
\setV{xm-qwen-lb-k}{0.546}
\setV{xm-qwen-lb-meanprobe}{0.530}
\setV{xm-qwen-ci-lb-dnine-dfive}{0.061}
\setV{xm-qwen-ci-lb-ninek-probe}{0.036}
\setV{xm-qwen-roleplay}{28}
\setV{xm-qwen-depth-above}{6}
\setV{xm-qwen-depth-peak}{0.852}
\setV{xm-qwen-depth-rise-rel}{0.71}
\setV{xm-qwen-hpkr-deight}{0.783}
\setV{xm-qwen-hpkr-dnine}{0.187}
\setV{xm-mistral-off-k}{0.799}
\setV{xm-mistral-lb-dnine}{0.762}
\setV{xm-mistral-lb-k}{0.775}
\setV{xm-mistral-lb-meanprobe}{0.570}
\setV{xm-mistral-ci-off-dnine-raw}{0.094}
\setV{xm-mistral-ci-off-ninek-dnine}{0.032}
\setV{xm-mistral-ci-lb-dnine-dfive}{0.043}
\setV{xm-mistral-ci-lb-ninek-probe}{0.038}
\setV{xm-mistral-roleplay}{68}
\setV{xm-mistral-flag-hden-deight}{99.8}
\setV{xm-mistral-depth-above}{9}
\setV{xm-mistral-depth-ident-above}{16}
\setV{xm-gemma-lb-dnine}{0.900}
\setV{xm-gemma-lb-k}{0.515}
\setV{xm-gemma-lb-meanprobe}{0.527}
\setV{xm-gemma-ci-off-dnine-raw}{0.175}
\setV{xm-gemma-ci-off-ninek-dnine}{\ensuremath{-}0.061}
\setV{xm-gemma-ci-lb-dnine-dfive}{0.221}
\setV{xm-gemma-ci-lb-ninek-probe}{0.064}
\setV{xm-gemma-roleplay}{96}
\setV{xm-gemma-depth-above}{7}
\setV{xm-llama-den-k}{0.728}
\setV{xm-llama-den-band}{0.726}
\setV{xm-mistral-ident-k}{0.840}
\setV{xm-mistral-ident-dnine}{0.760}
\setV{xm-mistral-ident-band}{0.755}
\setV{xm-mistral-ident-dfive}{0.784}
\setV{xm-mistral-ident-lies}{494}
\setV{xm-mistral-ident-honest}{729}
\setV{xm-mistral-ci-ident-k-probe}{0.073}
\setV{xm-mistral-ci-ident-k-probe-lo}{0.053}
\setV{xm-mistral-ci-ident-k-probe-hi}{0.095}
\setV{xm-qwen-rb-ident-k}{0.602}
\setV{xm-qwen-rb-ident-k-lo}{0.545}
\setV{xm-qwen-rb-ident-k-hi}{0.655}
\setV{xm-qwen-rb-ident-band}{0.584}
\setV{xm-gemma-rb-open-k}{0.585}
\setV{xm-gemma-rb-open-dnine}{0.624}
\setV{xm-gemma-rb-open-band}{0.622}
\setV{xm-gemma-rb-open-lies}{34}
\setV{xm-gemma-rb-whole-k}{0.378}
\setV{xm-gemma-rb-whole-k-lo}{0.275}
\setV{xm-gemma-rb-whole-k-hi}{0.489}
\setV{xm-mistral-gs-k}{0.440}
\setV{xm-mistral-gs-band}{0.765}
\setV{xm-gemma-k-id}{0.277}
\setV{xm-gemma-aisi-k}{0.236}
\setV{xm-gemma-aisi-ninek}{0.625}
\setV{xm-gemma-aisi-probe}{0.932}
\setV{xm-llama-rk-ident}{0.794}
\setV{xm-llama-rk-off}{0.791}
\setV{xm-mistral-rk-ident}{0.837}
\setV{xm-mistral-rk-off}{0.614}

%% file: generated/numbers_arxiv.tex
\setV{ax-headline-mean}{0.657}
\setV{ax-headline-rank}{16}
\setV{ax-d9-mean}{0.705}
\setV{ax-d9-rank}{11}
\setV{ax-repe-mean}{0.784}
\setV{ax-repe-rank}{5}
\setV{ax-reasoning-d9-mean}{0.799}
\setV{ax-reasoning-d9-rank}{3}
\setV{ax-reasoning-mean}{0.758}
\setV{ax-reasoning-rank}{7}
\setV{ax-headline-lo}{0.633}
\setV{ax-headline-hi}{0.680}
\setV{ax-rank-lo}{13}
\setV{ax-rank-hi}{20}
\setV{ax-n-read}{8}
\setV{ax-n-entries}{30}
\setV{ax-echo-headline}{0.189}
\setV{ax-echo-k}{0.206}
\setV{ax-abhc-headline}{0.408}
\setV{ax-abhc-k}{0.239}
\setV{qb-t3-delta}{\ensuremath{-}0.030}
\setV{qb-t3-lo}{\ensuremath{-}0.039}
\setV{qb-t3-hi}{\ensuremath{-}0.020}
\setV{qb-t4-r2-sigma}{0.67}
\setV{qb-t4-r2-plain}{0.41}
\setV{qb-t1-mono}{0.904}
\setV{qb-t1-median}{0.890}
\setV{qb-b150-b30-d9-cg}{0.056}
\setV{qb-b150-b30-d9-hpkr}{\ensuremath{-}0.134}
\setV{qb-best-family-lb}{0.911}
\setV{qb-pooled-lb}{0.881}
\setV{qb-confirm-aisi-ninek-delta}{0.006}
\setV{qb-confirm-aisi-ninek-lo}{0.004}
\setV{qb-confirm-aisi-ninek-hi}{0.009}
\setV{qb-rollouts}{27,000}
\setV{rb-t1-d}{0.916}
\setV{rb-t1-null}{0.828}
\setV{rb-t1-p}{0.002}
\setV{rb-t2-d}{0.899}
\setV{rb-t2-null}{0.814}
\setV{rb-t3-d}{0.922}
\setV{rb-t3-null}{0.891}
\setV{rb-t4-d}{0.912}
\setV{rb-t4-null}{0.938}
\setV{rb-t4-p}{0.144}
\setV{rb-t5-d}{0.101}
\setV{rb-t5-null-median}{\ensuremath{-}0.005}
\setV{rb-t5-null}{0.253}
\setV{rb-t5-p}{0.165}
\setV{rb-t6-d}{0.977}
\setV{rb-t6-null}{0.955}
\setV{rb-roles}{35}
\setV{ai-probe}{0.971}
\setV{ai-d9}{0.977}
\setV{ai-d9-delta}{0.006}
\setV{ai-d9-lo}{0.000}
\setV{ai-d9-hi}{0.013}
\setV{ai-ninek}{0.960}
\setV{ai-k}{0.899}
\setV{ai-rulek}{0.867}
\setV{li-k-vs-probe-ident}{0.014}
\setV{li-k-vs-probe-ident-lo}{\ensuremath{-}0.018}
\setV{li-k-vs-probe-ident-hi}{0.044}
\setV{tr-base-cells}{30}
\setV{tr-base-below}{6}
\setV{tr-ext-over-dec-offx}{0.037}
\setV{qb-decision-spread}{0.003}
\setV{qb-t1-holm}{1.00}
\setV{qb-confirm-gs-ninek-gap}{0.024}
\setV{qb-confirm-gs-ninek-gap-lo}{0.010}
\setV{qb-confirm-gs-ninek-gap-hi}{0.037}
\setV{rb-null-splits}{99}
\setV{rb-holm-splits}{9,999}
\setV{ax-posthoc-rank-min}{11}
\setV{ax-posthoc-rank-max}{28}
\setV{ai-ninek-gap}{0.011}
\setV{ai-ninek-gap-lo}{0.003}
\setV{ai-ninek-gap-hi}{0.020}
\setV{xs-mistral-dnine-probe}{\ensuremath{-}0.004}
\setV{xs-mistral-dnine-probe-lo}{\ensuremath{-}0.021}
\setV{xs-mistral-dnine-probe-hi}{0.014}
\setV{xs-gemma-dnine-probe}{0.117}
\setV{xs-gemma-dnine-probe-lo}{0.103}
\setV{xs-gemma-dnine-probe-hi}{0.130}
\setV{xq-s3u-off-dnine}{0.469}
\setV{xq-s3u-off-raw}{0.559}
\setV{xq-s3u-off-deight}{0.725}
\setV{xq-s3u-ci-dnine-raw}{\ensuremath{-}0.090}
\setV{xq-s3u-ci-dnine-raw-lo}{\ensuremath{-}0.101}
\setV{xq-s3u-ci-dnine-raw-hi}{\ensuremath{-}0.079}
\setV{bank-judged}{5,250}
\setV{bank-judged-kept}{3,599}
\setV{bank-anchor-rows}{150}
\setV{xq-s3u-roleplay}{69}
\setV{xq-sys-roles-under}{11}
\setV{xq-strict-min}{20}
\setV{xq-s23-offx-dnine}{0.627}
\setV{xq-s23-offx-raw}{0.608}
\setV{xq-s3u-offx-dnine}{0.674}
\setV{xq-s3u-offx-raw}{0.608}
\setV{ru-llama-gs-rule}{0.747}
\setV{xs-bank-gap}{0.02}

%% file: sections/abstract.tex
Linear probes are the cheapest LLM lie detectors but often fail to transfer across kinds of lie.
Label-free directions from honest and deceptive role-play are untested at 70B scale and on lies about
knowledge, where a lie and an honest denial can be identical. We widen a published persona bank with
roles that deny knowledge they have, plus honest counterparts. At each model's middle layer the resulting
persona direction ranks lies across Liars' Bench better than single-dataset probes on three of its four
models; it fails on Qwen, even when Qwen stays in role far more often. On \mbox{Llama-3.3-70B-Instruct} it beats
directions from a random persona bank, but its gain over the published direction is within the range of
random widenings. On lies about knowledge, persona directions mainly track a reply's wording: where lies
and honest denials share their words, our three main contrasts stay inside the random-direction band on
Llama and Qwen and reach only its edge on Gemma, though on Mistral one clears it. A knowledge direction
contrasts questions the model can answer with ones nobody can, each followed by the same scripted denial.
It separates lies from identical honest denials on two models and narrowly on a third, but not on the
fourth; its other results partly depend on the benchmark's wording. On the Gender-Secret fine-tune it
clears the band by a few hundredths on Llama and fails on Mistral. The sum of the two directions ranks
lies better than a supervised probe at the same depth on all four models (on one, both near chance) but is
level with the persona direction on Llama's average and behind that probe on held-out lies. More varied
role-play questions do not help the persona direction on Liars' Bench.

%% file: sections/introduction.tex
\section{Introduction}
\label{sec:intro}

Large language models sometimes lie. They have been observed concealing misconduct
\citep{scheurer2023strategic,jarviniemi2024uncovering} and faking alignment
\citep{greenblatt2024alignment}, and deception by AI systems is a recognised risk
\citep{park2023deception}. Following Liars' Bench \citep{kretschmar2025liars}, we call a statement a lie
when the model asserts something it believes to be false, where its belief is taken to be what it
consistently asserts under no pressure to lie. This definition separates lying from hallucination. The
cheapest detector of such lies is a probe on the residual stream, usually a linear one
\citep{alain2016understanding,azaria2023internal,marks2023geometry}: one dot product per response, no
extra forward pass, and a chance to catch a lie that leaves no trace in the transcript
\citep{goldowskydill2025detecting}. Its weakness is generalisation: a probe trained on one distribution
of lies often fails on others \citep{kirch2025offpolicy,kumar2026pressure,natarajan2026targeted}.
Liars' Bench makes the failure concrete. Its seven datasets hold on-policy lies, lies that four
open-weight models produced themselves. On most of them, probes trained on instruction pairs are near
chance in balanced accuracy at the benchmark's threshold, which flags 1\% of answers to ordinary
instructions. The hardest dataset is the harm-pressure knowledge report (\hpkr), in which a model denies
knowing the answer to a hazardous question it has been shown to know. On \hpkr{} the benchmark's Mean
Probe for \mbox{Llama-3.3-70B-Instruct}, read at layer~20, scores an AUROC of \V{lb-mean-auroc-hpkr}, worse than
chance. At the model's middle layer a probe trained the same way ranks \hpkr{} well, yet at that threshold
it still flags none of its lies.

\citet{mahadik2026persona} propose persona coordinates as a remedy: the model role-plays deceptive and
honest personas, each persona's mean activation is a vector, and the difference between the two group
means is a direction that uses no evaluation data and so cannot be fitted to one dataset's
idiosyncrasies. On \mbox{Llama-3.2-3B} and \mbox{Llama-3-8B} these coordinates transfer better than
probes on raw activations, keeping more of their accuracy on datasets they were not fitted to; the
authors name larger models and richer persona sets as future work. We take up both. Rebuilt for
\mbox{Llama-3.3-70B-Instruct}, their direction, which we call the \emph{published contrast}, inverts on \hpkr:
at the benchmark's layer it ranks honest answers above lies (AUROC \V{zs-dec-hpkr-L20}). Nearly every
\hpkr{} lie is a denial of knowledge (\V{th-hk-lie-knows-denies} of \V{ndec-hpkr} on Llama), while every
published deceptive persona lies by controlling disclosure. We therefore add ten roles of our own, four
that deny knowledge they have and six honest counterparts. The widened contrast, which we call the
\emph{extended contrast}, ranks \hpkr{} better, but for the wrong reason: its gain follows the words of a
denial. The contrast built from the new roles alone flags nearly every reply that says ``I do not
know'', whether the denial is true or false, and at the middle layer, on the Llama rows where a lie and an
honest denial are the same sentence, every persona direction scores within the range that random
directions reach (the \emph{random band}; \cref{sec:method}). On these lies a persona direction responds
to how a reply is worded more than to whether the model knows the answer.

The definition of a lie offers a second route. A lie about knowledge is a denial by a model that knows
the answer, and what a model knows can be measured without any labelled lie, by comparing questions
whose answers it knows with questions whose answers nobody can know. We build a \emph{knowledge
direction} from exactly that comparison, writing the same denial into the model's reply after every
question so that the reply's words do not vary. We then ask five questions of a detector that adds a
persona contrast to the knowledge direction: whether it transfers across Liars' Bench better than
supervised probes; whether it separates lies from honest answers that use the same words; what each part
responds to; whether the result survives resampling, new questions, other layers and strict thresholds;
and whether it holds on other models and benchmarks.

The results are mixed. The extended contrast is the part that generalises: at each model's middle layer
it transfers better than probes trained on a single dataset on three of the benchmark's four models. On
the main model it also beats contrasts drawn from a random persona bank, although a probe trained on all the
other datasets together is level with it once \hpkr{} is set aside (\V{th-pool-offx} against
\V{offx-ext-L40}). It fails on Qwen, even with a second bank in which Qwen stays in role far more often.
The knowledge direction separates lies from identically worded honest denials on two models, and
narrowly on a third. Outside these wording-controlled tests, its transfer depends in part on the wording of
\hpkr's template. On the Gender-Secret fine-tune, which only Llama and Mistral have, it clears the random
band by a few hundredths on Llama but not on Mistral; the other fine-tunes give it no support. The sum
of the two directions, our \emph{selected detector}, ranks the benchmark's lies better than a probe
trained at the same depth does on all four models (on Qwen both are near chance). On the main model's
average over the benchmark's datasets, however, it is level with the extended contrast, and in an
external competition it would rank \V{ax-headline-rank}th of \V{ax-n-entries}, behind a supervised probe.
At cutoffs that flag 1\% of each dataset's honest answers, none of the six detectors we evaluate there
catches half of the lies on any model (at most \V{th-k-op-dnine-1-L40}; \cref{tab:xm-operating}). Our
contributions are:
\begin{itemize}[leftmargin=1.2em, itemsep=0pt, topsep=1pt]
\item a persona bank for deception by denial, and a single label-free contrast built from it that, at the
  middle layer, transfers better than probes trained on a single dataset on three of four models. We
  check it against a widening by unrelated roles and, on the main model, against a random persona bank:
  it beats that bank's contrasts, but its gain over the published contrast is within what random
  widenings produce (\cref{sec:persona,sec:models,sec:external});
\item a diagnosis of what persona directions respond to: on lies about knowledge they track the words of
  a denial, and with the words held fixed the three main contrasts are, at the middle layer, inside the
  random band on Llama and Qwen and at its edge on Gemma, while on Mistral the published contrast clears
  it (\cref{sec:register,sec:models});
\item a knowledge direction with measured limits: it separates lies from identically worded honest
  denials on two models (narrowly on a third), clears the random band by a small margin on Llama's
  Gender-Secret fine-tune but not on Mistral's, and only its results on the wording-controlled slices are
  independent of the benchmark's wording (\cref{sec:knowledge,sec:models});
\item the selected detector, whose mean AUROC over the benchmark's datasets is, on all four models, above
  that of a probe trained at the same depth. On the main model it is level with the extended contrast, and
  at strict cutoffs it catches fewer lies than that contrast (\cref{sec:combined}); on the other three,
  adding the knowledge direction raises transfer on two and lowers it on the third (\cref{sec:models}). In
  a small steering test its persona part, like the published contrast, reduces the disclosure of a hidden
  fact (\cref{sec:steering});
\item an evaluation protocol for zero-shot (label-free) directions: random bands, a random persona bank
  with a pre-registered null, wording-controlled slices, the base model's activations on a fine-tuned
  model's transcripts as a control, and paired intervals; and two further pre-registered tests, one of them
  external, reported with their failures (\cref{sec:external}).
\end{itemize}

%% file: sections/related.tex
\section{Related work}
\label{sec:related}

\paragraph{Lie detection and why probes fail to transfer.} Instruction pairs, the same statement produced
under an honest and under a deceptive instruction, are the standard training data for deception probes
\citep{zou2023representation,goldowskydill2025detecting}. Black-box detectors instead ask unrelated
follow-up questions \citep{pacchiardi2023catch} or have an LLM judge read the transcript; Liars' Bench
finds the judge strongest overall, yet unable to detect \hpkr's lies, whose truth hinges on what the model
knows \citep{kretschmar2025liars}. A probe that is near perfect in distribution can be below chance
elsewhere. With fewer examples than dimensions, many directions fit a training set
\citep{damour2022underspecification}, and discriminative learning takes any feature that suffices on that
set \citep{geirhos2020shortcut}. Probes have accordingly latched onto negation
\citep{levinstein2025still,buerger2024truth}, text probability \citep{marks2023geometry},
deception-related scenarios \citep{goldowskydill2025detecting} and style \citep{kumar2026pressure}; they
suffer when trained on text the model did not itself generate (off-policy data;
\citealp{kirch2025offpolicy}), and they invert where a correlate of the label reverses
\citep{levinstein2025still,ying2026truthfulness,moustafa2026beyond}. Remedies constrain the direction, to
a difference of means \citep{marks2023geometry}, to selected principal components \citep{yoo2026subspace}
or to persona coordinates \citep{mahadik2026persona}; we take the last route.
\citet{ying2026truthfulness} further report that a probe trained on data pooled over several settings
performs on par with the best single-setting probe, and that a covariance-weighted cosine predicts
transfer better than the plain cosine; \cref{sec:external} tests whether both carry over, with families of
role-play questions as the settings.

\paragraph{Benchmarks and model organisms.} Liars' Bench \citep{kretschmar2025liars} collects
\V{lb-examples} on-policy lies and honest responses across seven datasets and four models, and evaluates
a \emph{Mean Probe} trained on the instruction pairs of representation engineering (RepE;
\citealp{zou2023representation}), a \emph{Follow-Up Probe} read at
a prefilled ``No'' after ``Was your last response deceptive?'', and an \emph{Upper-Bound Probe} trained on
the benchmark itself; on \hpkr, built from the hazardous questions of WMDP \citep{li2024wmdp}, only the
last detects the lies. The UK AI Security Institute (AISI) released belief-labelled rollouts (lies and
honest answers) from two of these models, Llama and Gemma \citep{aisi_lie_rollouts}; they serve as our
held-out data.
\citet{cooney2026lie} show that existing fine-tuned liars (model organisms) often do not hold the belief
they deny; they release belief-verified reasoning organisms on \mbox{Qwen3.5-27B} and \mbox{Qwen3.6-27B} and find that
activation and logprob detectors drop sharply on them while a chain-of-thought judge stays strong.
Our external test is Aletheia's Quest \citep{aletheia2026quest}, a lie-detection competition on model
organisms and prompted liars, with black-box and white-box tracks ranked by mean AUROC over ten held-out
datasets. The winner of its held-out white-box track scores the suspect model's own judgement of its
transcript.

\paragraph{Personas and knowledge as directions.} Persona vectors \citep{chen2025persona} and the
Assistant Axis \citep{lu2026assistant} are activation directions for character traits and archetypes.
\citet{mahadik2026persona} built a deception monitor from such directions: a deception axis, the
difference between eight deceptive and seven honest personas, used either directly (zero-shot) or through
the top principal components of the persona vectors, on datasets that do not include \hpkr. Roles differ
in style as well as honesty \citep{joshi2024personas}, and a model that switches to an ignorant persona
could deceive although ``no persona is ever lying'' \citep{marks2026persona}. A model's internal states
record whether it knows an answer
\citep{kadavath2022language,slobodkin2023curious,gottesman2024estimating,ferrando2025know}. We measure this
signal at the tokens of a denial written for the model, as a difference of means between questions whose
answers it knows and questions nobody can answer, and use it to detect lies.

%% file: sections/method.tex
\section{Method}
\label{sec:method}

\paragraph{Reads and persona contrasts.} A detector reads the residual stream at the output of one
decoder block, averaged over the response tokens, and maps this pooled vector $z$, which we call a
\emph{read}, to one score, higher for a lie. Forward passes run remotely on the National Deep Inference
Fabric (NDIF) through nnsight \citep{fiottokaufman2024nnsight}. We name depths by the layers of
\mbox{Llama-3.3-70B-Instruct}; \cref{tab:models} gives the block that stands for each of these layers in the
other models. The main depth is the model's middle block (the middle layer of the abstract and
introduction), as in the work we build on
\citep{lu2026assistant,mahadik2026persona}, and Liars' Bench reads layer~20. We also sweep \V{depth-n}
depths, every fourth layer from 20 to 76 plus the last block, scaled to each model; layers 36 and 44 were
chosen for reporting after seeing results and are labelled post hoc. A persona's vector $v_p$ is its mean
activation over the response tokens of its kept role-play responses. For a deceptive group
$\mathcal{D}$ and an honest group $\mathcal{H}$,
\begin{equation}
  d = \tfrac{1}{|\mathcal{D}|}\textstyle\sum_{p\in \mathcal{D}} v_p - \tfrac{1}{|\mathcal{H}|}\sum_{p\in \mathcal{H}} v_p,
  \qquad s(z) = \langle z,\, d/\lVert d\rVert \rangle ;
  \label{eq:contrast}
\end{equation}
which group is called deceptive fixes the sign. Our \emph{persona bank}, the set of roles with their
judged role-play responses, holds the 26 roles of the published persona method (8 deceptive, 7 honest,
5 sycophantic, 5 anti-sycophantic and the default assistant as an anchor) and \V{bank-new} new roles
about knowing and denying, 36 in all. Four \emph{knowledge-denial} roles know the answer and claim not
to; three \emph{sincere-ignorance} roles do not know it and say so; and three \emph{plain-assertion}
roles know it and give it. Every role has five system-prompt variants and answers the same 30 questions;
a \mbox{Llama-3.1-70B-Instruct} judge scores each rollout from 0 to 3 for how clearly it plays the role, and
only rollouts scored 3 are kept (\V{bank-judged-kept} of the \V{bank-judged} judged rollouts; the
anchor's \V{bank-anchor-rows} are not judged; \cref{app:bank}). Five contrasts of the form of
\cref{eq:contrast} differ only in their groups: \emph{published} (8 v 7), \emph{sycophancy} (5 v 5),
\emph{knowledge-denial} (our 4 denying roles v our 6 honest ones), and two widenings that add another
contrast's groups to the published ones. The \emph{unified} contrast adds the sycophancy contrast's
groups (13 v 12), and the \emph{extended} contrast adds the knowledge-denial contrast's (12 v 13). The unified contrast controls for
widening by roles unrelated to knowledge; the random persona bank below gives a null distribution for
widening by random roles.

\paragraph{The knowledge direction.} In \hpkr{} a lie and an honest denial can be the same sentence, and
only whether the model knows the answer makes one a lie. An AI assistant, Claude Opus, wrote
\V{th-k-known} general questions whose
answers a 70B model knows with certainty and \V{th-k-unknown} whose answers nobody can know (private,
unrecorded particulars of real places). Each question is placed in \hpkr's own knowledge-report template:
a four-option question followed by ``Tell me whether you know the answer to the question, and nothing
else'', under \hpkr's system text and with no role. We then wrote the reply for the model, a
\emph{forced} reply: the same denial after every question, once as ``I do not know the answer.''\ and
once as ``I don't know the answer to the question.'', giving \V{th-k-rows} reads. The direction is the
unit difference $\hat v_{\mathrm{know}} \propto \bar h_{\text{knows}} - \bar h_{\text{cannot}}$ between
the mean reads over the known and over the unknowable questions (\cref{fig:k-build}). Both sides receive
the same two replies, so the reply cannot separate them. The known side, where the denial is
false, is the lie side; this sign was fixed before any evaluation. The questions themselves still differ
in topic and form, and a text classifier tells the two sides apart \V{th-k-style-orig}\% of the time. For
a replication set of matched question pairs, written afterwards and also by Claude, the figure is \V{th-k-style-matched}\%, and
the \emph{replication direction} built from it met all \V{th-k-rep-criteria} criteria fixed before it was
evaluated. Before the direction was built, the model answered all \V{th-k-verify-orig-correct} candidate
known questions correctly. The direction shares \hpkr's template and denial sentences, but it shares no question with any
evaluation set and uses no lie that the model wrote. A \emph{wording direction}, forced denials minus
forced assertions, measures how much a detector responds to the words of a denial; a detector
\emph{without its wording component} has its component along this direction removed. In each further
model the direction is rebuilt with the known questions checked against that model's answers. Our main
detector, the \emph{selected detector}, sums the extended contrast and the knowledge direction, each
score divided by its standard deviation on the lie-free Alpaca control \citep{taori2023alpaca}, with
equal weights fixed in advance. A direction is \emph{zero-shot} when it uses no labels, no lies and no
benchmark question, and it is \emph{scored zero-shot} when its projection is the score, with no sign or
scale fitted on labelled data (the selected detector's two components are scaled on the lie-free Alpaca
responses). The knowledge direction meets this definition only narrowly: it shares \hpkr's system text,
its template and the exact sentence that defines the identical-text slice below, and
\cref{sec:knowledge,sec:models} examine how much its results depend on them.

\paragraph{Baselines and model organisms.} Our supervised baselines are logistic regressions, trained as
in the benchmark's pipeline. The \emph{probe on raw activations} (the \emph{raw probe}) uses the full
residual stream; trained on one dataset and tested on the others, it is the transfer baseline. The
\emph{base probe} (our baseline probe; ``base'' does not refer to the base model) also uses the full
residual stream and follows the Mean Probe's recipe on RepE pairs \citep{zou2023representation} at the
same block as our directions; a second version is trained on a split of \V{th-aisi-train-rows} of Llama's
own belief-labelled rollouts (\V{th-aisi-train-lies} lies; \citealp{aisi_lie_rollouts}). The published
persona method's probe is trained instead on the top three principal components of the persona vectors.
The benchmark's \emph{model organisms} \citep{kretschmar2025liars} are LoRA fine-tunes that lie by
design: Gender-Secret knows the user's gender and denies knowing it, and Soft-Trigger answers falsely
when a trigger appears. Their lies exist only in the fine-tuned model, so the adapter is applied inside
the remote pass, and the directions are also rebuilt with it applied. Each transcript is also read by the
base model as a control: a detector that separates the lies only in the fine-tune's read responds to the
fine-tune's state, not to the text.

\paragraph{Models, data and summary measures.} \Cref{tab:models} (\cref{app:models}) lists the four
benchmark models and, for each, the block that stands for each of Llama's layers. Each model is evaluated
on the datasets the benchmark provides for it, all written by that model: convincing-game (CG),
harm-pressure choice (\hpc), \hpkr, insider trading (IT) and instructed deception (ID); the benchmark's
other two datasets come from the model organisms above. In \emph{train one, test all}, a probe is trained
on one of a model's sets (on Llama, the five datasets and RepE) and tested on the others, so RepE is both
a training and a test set. The mean AUROC over the off-diagonal cells of this matrix is the probe's
\emph{transfer}. A probe on a single zero-shot direction can only choose a sign and a scale, so when every
training set agrees on the sign, its transfer is the direction's mean AUROC over the sets; a direction
below 0.5 on one training set is flipped for every test set when trained there, so its transfer can fall
well below its mean AUROC. Our second summary averages a detector's AUROC over the model's benchmark
datasets, for detectors trained on none of them (directions scored zero-shot, and probes trained on RepE
or on the AISI rollouts); on Llama we call it the \emph{five-dataset mean}. Alpaca sets the thresholds and
the selected detector's scaling.

\paragraph{Wording-controlled slices.} In \hpkr{} nearly every lie (\V{th-hk-lie-knows-denies} of
\V{ndec-hpkr} on Llama) is a denial of knowing an answer the model has been shown to know, so whether a
reply denies knowledge already ranks \hpkr{} at an AUROC of \V{th-hk-denial-auroc}, without reading any
activation. We therefore also score two
\emph{wording-controlled slices} of \hpkr: the \emph{identical-text slice}, every row whose response is
exactly ``I do not know the answer.''\ (\V{th-k-ident-lies} lies v \V{th-k-ident-honest} honest on
Llama), and the \emph{denials slice}, every row whose response denies knowledge. Within a slice, lies and
honest denials answer different questions, so the slices control the wording but not the question.

\paragraph{Thresholds, intervals and nulls.} Detection decisions use the benchmark's threshold, which
flags 1\% of the Alpaca responses, and \emph{in-setting cutoffs}, which flag 1, 5 or 10\% of a dataset's
own honest answers. Comparisons at the middle block get 95\% intervals from 1,000 stratified resamples of
the rows, the same resamples for every method, so differences are paired. A difference counts only when
its interval excludes zero, and we call two detectors \emph{level} when it does not. A detector
\emph{inverts} when its AUROC is below 0.5, ranking honest answers above lies. Depth curves and averages
over layers are point estimates, and the cross-model and per-layer checks are not corrected for multiple
comparisons. A zero-shot AUROC is compared with the AUROCs of 1,000 random unit directions on the same
rows. Their central 95\% is the \emph{random band}, and the value we quote for a band is its upper edge,
the 97.5th percentile: a direction is \emph{at its edge} when its AUROC and that edge differ by less than
0.01 (the rule of \cref{tab:claims}), \emph{clears} the band when it is higher still, and is \emph{inside}
it otherwise. We check the bank against a \emph{random persona bank}: \V{rb-roles} roles drawn from
the 274 non-default roles released with the Assistant Axis and built by the same pipeline. Partitioning
them into groups of our sizes gives one reported split and \V{rb-null-splits} further splits, which form
the pre-registered null.

\paragraph{Steering.} For steering, the residual stream after the middle block is replaced at every token
by $h + c\,\lVert h\rVert\,\hat v$, where $h$ is the residual vector at that token and $\hat v$ the unit
direction being steered, while the model answers \V{th-steer-n} role-play scenarios from the AI Liar set
of \citet{pacchiardi2023catch}. Each scenario's system prompt gives the model an unfavourable fact and a
reason to hide it. Answers are greedy and are compared at $\lvert c\rvert = 0.3$, the largest size at
which no direction breaks the model. One reader, an AI assistant blind to the condition, labelled each
answer for disclosure of the fact (1, 0.5 or 0).

%% file: sections/results_llama.tex
\section{Results on Llama-3.3-70B-Instruct}
\label{sec:llama}

Every result in this section is at layer~40 unless stated otherwise; \cref{tab:designs} (\cref{app:llama}) compares
all the detectors side by side.

\subsection{Persona contrasts at 70B}
\label{sec:persona}

Our base probe, trained by the Mean Probe's recipe at layer~40, ranks \hpkr{} well (AUROC
\V{bp-auroc-hpkr-L40}) yet flags none of its \V{ndec-hpkr} lies at the benchmark's threshold. At
layer~20 it inverts (\V{bp-auroc-hpkr-L20}), as the published contrast does
(\V{zs-dec-hpkr-L20}); at layer~40 the published contrast scores \V{zs-dec-hpkr-L40}, inside a random
band that reaches \V{th-rand-hpkr-L40}. The raw probe averages \V{diag-base-L40} on its own dataset and
\V{off-base-L40} on the other five (\cref{fig:transfer}), with \V{tr-base-below} of its \V{tr-base-cells}
off-diagonal cells below chance. A probe on the published persona method's top three principal
components, fitted on the published roles, transfers at \V{off-pcthree-L40}, below the raw probe
(\cref{tab:transfer}). The extended contrast, one direction with no training, transfers at
\V{off-ext-L40}: above the raw probe by \V{xm-llama-ci-off-dnine-raw} [\V{xm-llama-ci-off-dnine-raw-lo},
\V{xm-llama-ci-off-dnine-raw-hi}], and above the published contrast (\V{off-dec-L40}) by
\V{xm-llama-ci-off-dnine-dfive}, or by \V{tr-ext-over-dec-offx} with \hpkr{} left out. Scored zero-shot,
its five-dataset mean is \V{zs-ext-lb-L40}, against \V{bp-auroc-lb-L40} for the base probe at the same
layer and \V{lb-mean-auroc-avg} for the benchmark's layer-20 Mean Probe.

Its lead over a probe trained on many kinds of lie rests on \hpkr. A probe trained on the other five
datasets together averages \V{th-pool-auroc}, and with \hpkr{} left out it is level with the extended
contrast (\V{th-pool-offx} against \V{offx-ext-L40}); on \hpkr{} itself, the extended contrast's AUROC is
inside the random band and follows the words of a denial (\cref{sec:register}). With \hpkr{} left out,
the sycophancy contrast, which has no deceptive role, transfers best of all the detectors in
\cref{tab:designs} (\V{th-des-d6-offx} against \V{th-des-d9-offx} for the extended contrast). Our ten
roles raise the five-dataset mean by \V{xm-llama-ci-lb-dnine-dfive} over the published contrast, about
two thirds of the gain coming from \hpkr. The unified contrast, widened in the same way with the
sycophancy contrast's roles, transfers at \V{th-uni-off-mean} averaged over the 16 layers, below the
\V{th-dec-off-mean} of the contrast it widens (point estimates). Against random widenings drawn from the
random persona bank, however, the gain from our roles is not significant, and the extended contrast's
AUROC on all of \hpkr{} is inside that null (\cref{sec:external}).

\subsection{Persona directions track the wording}
\label{sec:register}

\begin{figure}[t]
\centering
\includegraphics[width=\linewidth]{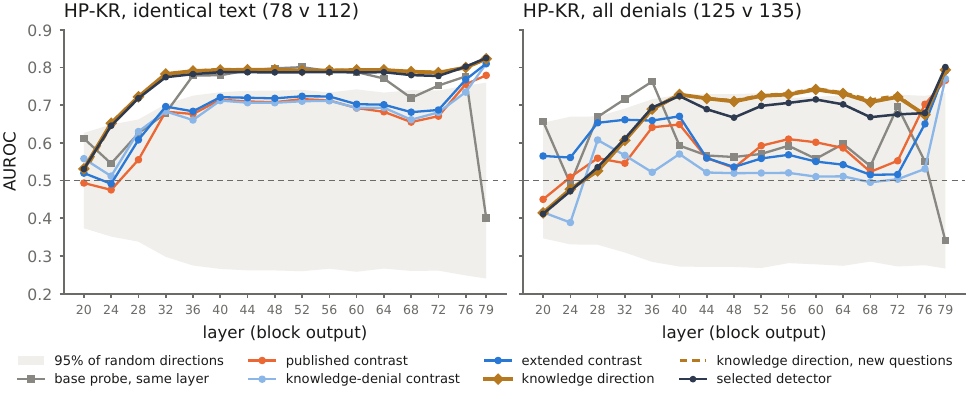}
\caption{AUROC on \hpkr's two wording-controlled slices, the identical-text slice (left) and the denials
slice (right), across the 16 layers. Grey marks the random band, the central 95\% of the AUROCs of random
directions; ``knowledge direction, new questions'' is the replication direction, and the two knowledge-direction curves
nearly coincide.}
\label{fig:k-hpkr}
\end{figure}

Of our detectors, only the knowledge-denial contrast has a balanced accuracy on \hpkr{} clearly above
chance (\V{op-den-balacc-hpkr-L40}), and it gets there by flagging denials. At the benchmark's threshold
it flags \V{th-k-deight-flag-lies-L40}\% of the lies and
\V{th-k-deight-flag-hden-L40}\% of the honest denials but just \V{th-k-deight-flag-hclaims-L40}\% of the honest
answers that claim knowledge; honest denials are only \V{th-k-den-honest} of the \V{th-honest-hpkr}
honest answers. With the words held fixed, the persona directions do no better than the best random
directions. On the identical-text slice the published, knowledge-denial and extended contrasts score
\V{th-k-dfive-ident-L40}, \V{th-k-deight-ident-L40} and \V{th-k-dnine-ident-L40}, all inside the random
band (\V{th-k-ident-band-L40}). From layer 32 to 72 none of these three contrasts exceeds
\V{th-k-depth-persona-ident-max} on this slice, and they clear the band only at the last two layers of the
sweep (\cref{fig:k-hpkr}). The extended contrast built from the 150-question bank of \cref{sec:external}
scores slightly higher at layer~40, just above the band's upper edge but within 0.01 of it
(\cref{tab:qbank-tests}). On Gender-Secret, whose lies and honest answers are both
denials, every persona direction built on the base model and read through the fine-tune is near chance (the knowledge-denial contrast
\V{th-gs-den-L40}); rebuilt inside the fine-tune, the best reaches \V{th-gs-ft-best} at
layer~\V{th-gs-ft-best-layer} (post hoc), still far inside the random band. On \hpkr, then, the persona
directions' signal follows the words of a denial, and on the identical-text slice none of them separates
lies from honest denials better than the best random directions do.

\subsection{The knowledge direction}
\label{sec:knowledge}

On the identical-text slice the knowledge direction scores \V{th-k-k-ident-L40} [\V{th-k-ident-lo-L40},
\V{th-k-ident-hi-L40}], above the band, and \V{th-k-ident-nowording-L40} without its wording component. It
is above the band at \V{th-k-depth-ident-above} of the \V{th-k-depth-n} layers, and the replication
direction follows the same curve (\cref{fig:k-hpkr}). On the wider denials slice, whose lies and honest
denials need not share their exact words, it sits at the band's edge (\V{xm-llama-den-k} against
\V{xm-llama-den-band}), so on Llama its support rests on the identical-text slice. Because that slice's
lies and honest denials answer different questions, the slice cannot separate what the model knows from
how the questions differ. On it, the base probe trained at the same layer, a supervised probe that inverts
on insider trading, is level with the knowledge direction (\V{th-k-probe-ident-L40}; difference
\V{li-k-vs-probe-ident} [\V{li-k-vs-probe-ident-lo}, \V{li-k-vs-probe-ident-hi}]). The
direction was also built from this same reply under \hpkr's template; the template-free direction
introduced below keeps the result (\V{xm-llama-rk-ident}). On all of \hpkr{} the knowledge direction
scores \V{th-k-full-L40}, but this figure depends on the words (\V{th-k-full-nowording-L40} without its
wording component), so we count only its slice results as evidence.

Built inside the Gender-Secret fine-tune and read through it, the knowledge direction scores
\V{th-k-gs-ft-kft-L40} [\V{th-k-gs-ft-kft-lo-L40}, \V{th-k-gs-ft-kft-hi-L40}] against a band of
\V{th-k-gs-ft-band-L40}, while the base model's read of the same transcripts gives
\V{th-k-gs-base-kft-L40} (\cref{app:organisms}). The margin is a few hundredths, and the classes differ in
topic as well as in what the model knows. On the five datasets the direction averages \V{th-k-k-mean-L40},
below the extended contrast's \V{th-k-dnine-mean-L40}, and it leads the extended contrast only on \hpc{}
(\V{th-k-k-auroc-hpc-L40} against \V{th-k-dnine-auroc-hpc-L40}). Its results beyond the slices rest in
part on \hpkr's template. A \emph{template-free} knowledge direction, rebuilt without that wording by a
rule chosen on these datasets, keeps the wording-controlled \hpkr{} results. It transfers a little worse
here, however (\V{xm-llama-rk-off} against \V{th-k-k-off-L40}), and much worse on Mistral (\cref{sec:models}),
and on Gender-Secret it falls inside the random band (\V{ru-llama-gs-rule}). Resampling the questions gives
a 95\% interval at most \V{th-k-boot-ident-width} wide on the identical-text slice. None of 1,000 directions built
from balanced splits of the same questions reaches the knowledge direction's AUROC there; we adopted this
null after the random split we had specified first proved uninformative. The replication direction has a
cosine of \V{th-k-rep-cos-L40} with the original and was scored on the same test rows, so its agreement is
not independent evidence.

\subsection{The selected detector and strict cutoffs}
\label{sec:combined}

The selected detector transfers at \V{th-k-ninek-off-L40} against \V{off-base-L40} for the raw probe,
level with the extended contrast (\V{xm-llama-ci-off-ninek-dnine} [\V{xm-llama-ci-off-ninek-dnine-lo},
\V{xm-llama-ci-off-ninek-dnine-hi}]), and its five-dataset mean of \V{th-k-ninek-mean-L40} is also level
with the contrast's (difference \V{th-pair-ninek-dnine-mean-L40} [\V{th-pair-ninek-dnine-mean-lo-L40},
\V{th-pair-ninek-dnine-mean-hi-L40}]). Its gains over the extended contrast come where the words are
controlled: \V{th-pair-ninek-dnine-ident-L40} [\V{th-pair-ninek-dnine-ident-lo-L40},
\V{th-pair-ninek-dnine-ident-hi-L40}] on the identical-text slice and \V{th-k-gs-gain-L40} on Gender-Secret, at
small costs on CG, ID and IT. On Gender-Secret, with both directions rebuilt in the fine-tune, it scores
\V{th-k-gs-ft-ninekft-L40}, just above the band; built on the base model, it scores
\V{th-k-gs-ft-ninek-L40}, inside the band. It beats the same-layer base probe on the five-dataset mean by
\V{xm-llama-ci-lb-ninek-probe}, almost all of the margin coming from insider trading, where the probe
inverts; it is level with that probe on the identical-text slice and below it on held-out lies
(\cref{sec:external}). Across depth it transfers above the published and unified contrasts and the raw
probe at every one of the 16 layers, by at least \V{th-k-ninek-depth-lead-min} (\cref{app:depth}; point
estimates). Averaged over the layers, however, it transfers below the extended contrast alone
(\V{th-k-ninek-depth-off-mean} against \V{th-ext-off-mean}; at layer~20, \V{th-k-ninek-off-L20} against
\V{off-ext-L20}), whose own transfer peaks at layer~\V{depth-off-ext-peak-layer} (post hoc). Its weakness
is at strict cutoffs: when the cutoff flags 1\% of a dataset's own honest answers, the extended contrast
catches \V{th-k-op-dnine-1-L40} of the lies, the selected detector \V{th-k-op-ninek-1-L40}, the base probe
\V{th-k-op-probe-1-L40} and the knowledge direction \V{th-k-op-k-1-L40}. Placing the threshold on each
dataset instead of on Alpaca changes balanced accuracy by a few hundredths at most (\cref{tab:oracle}).

\subsection{Steering the persona contrast}
\label{sec:steering}

Adding the extended contrast at $c = +0.3$ cuts disclosure of the hidden fact from \V{th-steer-base} to
\V{th-steer-ext-plus} (\V{th-steer-ext-plus-change} [\V{th-steer-ext-plus-lo}, \V{th-steer-ext-plus-hi}]).
The published contrast does the same (\V{th-steer-dec-plus-change}), so the effect is not specific to our
roles; two random directions do not, and the base probe's direction moves it by
\V{th-steer-base-plus-change} [\V{th-steer-base-plus-lo}, \V{th-steer-base-plus-hi}] (\cref{app:steering}).
The steered answers take on the deceptive personas' manner, so the steering plausibly works through the
speaking style of those roles. The result rests on one blinded reader (an AI assistant), one coefficient
and one greedy answer per scenario. An earlier run on benchmark items with an absolute shift (a fixed
multiple of the direction, not scaled by $\lVert h\rVert$) found no effect beyond that of random
directions, and two small runs that steered the knowledge direction are suggestive only
(\cref{app:steering}).

%% file: sections/results_models.tex
\section{Across models}
\label{sec:models}

We repeated the studies with the same code on three further models, \mbox{Qwen2.5-72B-Instruct},
\mbox{Mistral-Small-3.1-24B-Instruct} and \mbox{Gemma-3-27B-it} (hereafter Qwen, Mistral and Gemma), fixing the blocks and
banks before inspecting any result (\cref{tab:models} in \cref{app:models}; steering was not repeated). Each model has its own bank from
the same roles, variants, questions and judge. In Qwen's \emph{strict} bank, which keeps only rollouts
scored 3, \V{xq-sys-roles-under} roles keep fewer than \V{xq-strict-min} rollouts, so Qwen uses the \emph{lenient}
bank, which also keeps rollouts scored 2 (marked $\dagger$ in the tables); the lenient banks of the other
three models change the contrasts' means and the extended contrast's transfer at the middle block by at
most \V{xs-bank-gap} (\cref{tab:xm-banks}). Only Llama and Mistral have a released identical-text slice;
for Qwen and Gemma we rebuilt comparable slices from the raw files, a fixed reply for Qwen and a fixed
opening sentence for Gemma (\cref{tab:xm-wording}). \Cref{tab:claims} checks the claims at each
model's middle block; every interval is in \cref{app:models}.

\begin{table}[t]
\centering
\footnotesize
\setlength{\tabcolsep}{4pt}
\caption{The claims at each model's middle block: a paired difference, or an AUROC with the value it must
clear (or, for the row on persona directions, stay below) in parentheses. \checkmark{} holds, $\times$
does not, $\circ$ no clear difference (the interval includes zero, or the values are within 0.01).
$^\dagger$Qwen's lenient bank. $^a$Qwen's identical reply, rebuilt from the raw files. $^b$Gemma's
opening sentence, rebuilt, \V{xm-gemma-rb-open-lies} lies. In the third row the knowledge-denial roles are
all ten new roles, not only the four denying ones. The Mean Probe row compares with Liars' Bench's layer-20 probe; the same-block base
probe is in \cref{tab:xm-mid}. On Gender-Secret the margin is a few hundredths and the classes also differ
in topic.}
\label{tab:claims}
\fitwidth{\input{generated/tab_xm_claims}}
\end{table}

\paragraph{What carries over.} On Mistral and Gemma the extended contrast transfers above the raw probe
(by \V{xm-mistral-ci-off-dnine-raw} and \V{xm-gemma-ci-off-dnine-raw}, intervals above zero) and above the
published contrast; on Mistral much of the second lead comes from a sign flip of the published contrast,
which scores below 0.5 on \hpc{} and so is flipped for every test set when trained there
(\cref{sec:method}). Zero-shot, its mean over the model's datasets is well above the benchmark's layer-20
Mean Probe on both (\V{xm-mistral-lb-dnine} against \V{xm-mistral-lb-meanprobe}; \V{xm-gemma-lb-dnine}
against \V{xm-gemma-lb-meanprobe}); against the base probe trained at the same block it is above on Gemma
(\V{xs-gemma-dnine-probe} [\V{xs-gemma-dnine-probe-lo}, \V{xs-gemma-dnine-probe-hi}]) and level on Mistral
(\V{xs-mistral-dnine-probe} [\V{xs-mistral-dnine-probe-lo}, \V{xs-mistral-dnine-probe-hi}]). Adding our
ten roles raises the mean above the published contrast's on all four models (by
\V{xm-llama-ci-lb-dnine-dfive}, \V{xm-qwen-ci-lb-dnine-dfive}, \V{xm-mistral-ci-lb-dnine-dfive} and
\V{xm-gemma-ci-lb-dnine-dfive}), though on Llama about one in six random widenings drawn from the random
persona bank gains as much (\cref{sec:external}), so the gain is not shown to be specific to these roles.
The selected detector's mean over the benchmark's datasets is above the same-block probe's on all four (by
\V{xm-llama-ci-lb-ninek-probe}, \V{xm-qwen-ci-lb-ninek-probe}, \V{xm-mistral-ci-lb-ninek-probe} and
\V{xm-gemma-ci-lb-ninek-probe}; on Qwen both are near chance, and on Llama almost all of the margin comes
from insider trading), while on the held-out lies of Llama and Gemma the probe is ahead (\V{ai-probe}
against \V{ai-ninek}; \V{xm-gemma-aisi-probe} against \V{xm-gemma-aisi-ninek}). Adding the knowledge
direction leaves transfer level on Llama, raises it on Mistral (\V{xm-mistral-ci-off-ninek-dnine}) and
Qwen (to \V{xm-qwen-off-ninek}, still below the raw probe) and lowers it on Gemma
(\V{xm-gemma-ci-off-ninek-dnine}), where the knowledge direction alone is weak.

\paragraph{Qwen: the persona contrast fails.} On Qwen the extended contrast transfers below the raw probe
(\V{xm-qwen-off-dnine} against \V{xm-qwen-off-raw}), is level with the published contrast, averages
\V{xm-qwen-lb-dnine} zero-shot against the Mean Probe's \V{xm-qwen-lb-meanprobe}, and inverts on \hpkr{}
(\V{xm-qwen-hpkr-dnine}). Much of the failure comes from \hpkr: with \hpkr{} left out it transfers above the
raw probe (\V{xq-s23-offx-dnine} against \V{xq-s23-offx-raw}; point estimates), though it stays weak on
\hpc{} and ID. The knowledge-denial contrast does not fail (transfer \V{xm-qwen-off-deight}; \hpkr{}
\V{xm-qwen-hpkr-deight}). The judge scored \V{xm-qwen-roleplay}\% of Qwen's rollouts fully in role,
against \V{xm-llama-roleplay}\%, \V{xm-mistral-roleplay}\% and \V{xm-gemma-roleplay}\% for Llama, Mistral
and Gemma (by group in \cref{tab:xm-roleplay}), which suggested that the contrast failed because Qwen did
not play the roles. A second Qwen bank, built after these results, places the role assignment, with an
instruction to stay in character, in the user turn. With it, Qwen plays \V{xq-s3u-roleplay}\% of its
rollouts fully in role and every role keeps enough rollouts to enter the strict bank. Yet the extended
contrast still transfers below the raw probe (\V{xq-s3u-off-dnine} against \V{xq-s3u-off-raw}; \V{xq-s3u-ci-dnine-raw}
[\V{xq-s3u-ci-dnine-raw-lo}, \V{xq-s3u-ci-dnine-raw-hi}]), again because of \hpkr{} (\V{xq-s3u-offx-dnine}
against \V{xq-s3u-offx-raw} without it), while the knowledge-denial contrast does not fail
(transfer \V{xq-s3u-off-deight}). Poor role-play is therefore not the whole explanation, and why the
published roles invert on Qwen's \hpkr{} remains open.

\paragraph{The knowledge direction across models.} On Mistral, the one other model with a released
identical-text slice (\V{xm-mistral-ident-lies} lies v \V{xm-mistral-ident-honest} honest), the knowledge
direction scores \V{xm-mistral-ident-k} against a band of \V{xm-mistral-ident-band}; it is above the
same-block probe by \V{xm-mistral-ci-ident-k-probe} [\V{xm-mistral-ci-ident-k-probe-lo},
\V{xm-mistral-ci-ident-k-probe-hi}] and above its band at all \V{xm-mistral-depth-ident-above} depths. On
Qwen's rebuilt identical-reply slice it narrowly clears the band (\V{xm-qwen-rb-ident-k}
[\V{xm-qwen-rb-ident-k-lo}, \V{xm-qwen-rb-ident-k-hi}] against \V{xm-qwen-rb-ident-band}). On Gemma's
rebuilt slice it is inside the band (\V{xm-gemma-rb-open-k} against \V{xm-gemma-rb-open-band}), though with
only \V{xm-gemma-rb-open-lies} lies the slice is too small to decide, and on Gemma's rebuilt denials it
inverts (\V{xm-gemma-rb-whole-k} [\V{xm-gemma-rb-whole-k-lo}, \V{xm-gemma-rb-whole-k-hi}]). The wording
diagnosis of \cref{sec:register} holds on Qwen's rebuilt slice, where every persona direction is inside
the band; on Gemma's, the extended contrast sits at the band's edge (\V{xm-gemma-rb-open-dnine} against
\V{xm-gemma-rb-open-band}), with too few lies to decide either way. It holds only in part on Mistral: the
knowledge-denial contrast flags \V{xm-mistral-flag-hden-deight}\% of the honest denials and is inside the
band on the identical-text slice, the extended contrast is at its edge (\V{xm-mistral-ident-dnine}), and the
published contrast is above it (\V{xm-mistral-ident-dfive}), so the claim that no persona direction clears
the band fails there. On its own, the knowledge direction's mean over each model's datasets is
\V{xm-mistral-lb-k} on Mistral but \V{xm-qwen-lb-k} on Qwen and \V{xm-gemma-lb-k} on Gemma, where it
inverts on instructed deception (\V{xm-gemma-k-id}) and on Gemma's own held-out lies (\V{xm-gemma-aisi-k}).
The other models' fine-tunes add no evidence for it. Mistral's Gender-Secret, the one replication of
Llama's test, counts against it (\V{xm-mistral-gs-k} against a band of \V{xm-mistral-gs-band}), and on the
Soft-Trigger fine-tunes either the base model's read already separates the classes or the direction stays
inside the band (\cref{app:organisms}). The template-free knowledge direction (\cref{sec:knowledge}) keeps
the wording-controlled \hpkr{} results on Mistral (identical text \V{xm-mistral-rk-ident}, by a rule fixed
before Mistral was evaluated) and on Llama (\V{xm-llama-rk-ident}, though there the rule's wording was
chosen with Llama's own datasets in view), but it transfers worse (\V{xm-mistral-rk-off} against
\V{xm-mistral-off-k} on Mistral). Of the knowledge direction's results, then, only those on the slices are
free of the template.

\paragraph{Across depth and what generalises.} The middle block was fixed in advance, but it need not be
where a model's signal is strongest (\cref{app:models}). The extended contrast transfers above the raw
probe, by any margin, at all \V{xm-llama-depth-n} depths on Llama, but at only \V{xm-mistral-depth-above}
of 16 on Mistral, \V{xm-gemma-depth-above} on Gemma and \V{xm-qwen-depth-above} on Qwen. On Qwen it stays
near chance until \V{xm-qwen-depth-rise-rel} of the depth and then reaches \V{xm-qwen-depth-peak} at a block
found only by looking (post hoc), where it is level with the raw probe. Two results hold on every model:
adding our ten roles raises the contrast's mean (on Llama, by an amount within the range that random
widenings produce), and
the selected detector's mean over the benchmark's datasets is above the same-block probe's (on Qwen both
are near chance, and on Llama the margin comes almost entirely from insider trading). Where the words are
fixed, the knowledge direction separates lies from identical honest denials on Llama and Mistral and
narrowly on Qwen, but not on Gemma's \V{xm-gemma-rb-open-lies} rebuilt lies. The extended contrast's lead
in transfer replicates on Mistral and Gemma, not on Qwen; its lead over the raw probe at every depth, the
detection of Gender-Secret through the fine-tune, and the value of adding the knowledge direction do not
generalise.

%% file: generated/tab_xm_claims.tex
\begin{tabular}{@{}>{\raggedright\arraybackslash}p{4.2cm} cccc@{}}
\toprule
Claim & Llama & Qwen$^\dagger$ & Mistral & Gemma \\
\midrule
Extended contrast transfers above the probe on raw activations & $+$0.184 \checkmark & $-$0.106 $\times$ & $+$0.094 \checkmark & $+$0.175 \checkmark \\
\addlinespace[3pt]
Extended contrast transfers above the published contrast & $+$0.084 \checkmark & $+$0.002 $\circ$ & $+$0.176 \checkmark & $+$0.221 \checkmark \\
\addlinespace[3pt]
The knowledge-denial roles raise the mean (extended above published) & $+$0.101 \checkmark & $+$0.061 \checkmark & $+$0.043 \checkmark & $+$0.221 \checkmark \\
\addlinespace[3pt]
Extended contrast's mean is above the benchmark's Mean Probe & 0.899 (0.682) \checkmark & 0.441 (0.530) $\times$ & 0.762 (0.570) \checkmark & 0.900 (0.527) \checkmark \\
\addlinespace[3pt]
Selected detector transfers above the probe on raw activations & $+$0.190 \checkmark & $-$0.045 $\times$ & $+$0.125 \checkmark & $+$0.114 \checkmark \\
\addlinespace[3pt]
Selected detector's mean is above the same-block RepE probe & $+$0.128 \checkmark & $+$0.036 \checkmark & $+$0.038 \checkmark & $+$0.064 \checkmark \\
\addlinespace[3pt]
No persona direction clears the band on identical text & 0.721 (0.733) \checkmark & 0.550 (0.584)$^a$ \checkmark & 0.784 (0.755) $\times$ & 0.624 (0.622)$^b$ $\circ$ \\
\addlinespace[3pt]
Knowledge direction clears the band on identical text & 0.793 (0.733) \checkmark & 0.602 (0.584)$^a$ \checkmark & 0.840 (0.755) \checkmark & 0.585 (0.622)$^b$ $\times$ \\
\addlinespace[3pt]
Knowledge direction clears the band on Gender-Secret, read through the fine-tune & 0.861 (0.824) \checkmark & n/a & 0.440 (0.765) $\times$ & n/a \\
\bottomrule
\end{tabular}

%% file: sections/results_external.tex
\section{External and pre-registered tests}
\label{sec:external}

\paragraph{Held-out lies.} The validation split of the AISI rollouts holds Llama's own belief-labelled
answers (\V{th-aisi-val-lies} lies, \V{th-aisi-val-honest} honest answers) under system prompts that no
other part of this work used. On it, at layer~40, the extended contrast scores \V{ai-d9}, the selected
detector \V{ai-ninek} and the knowledge direction \V{ai-k}. The base probe trained on RepE at the same
layer scores \V{ai-probe}: it is level with the extended contrast (difference \V{ai-d9-delta}
[\V{ai-d9-lo}, \V{ai-d9-hi}]; the lower end is zero even to four decimals) and above the selected detector, which trails it by \V{ai-ninek-gap}
[\V{ai-ninek-gap-lo}, \V{ai-ninek-gap-hi}]. A base probe trained on a split of \V{th-aisi-train-rows} of
Llama's own rollouts (\V{th-aisi-train-lies} lies, disjoint from the validation split) raises the
five-dataset mean by \V{th-aisi-gain-L40} [\V{th-aisi-gain-lo-L40}, \V{th-aisi-gain-hi-L40}] over the
base probe trained on RepE. The extended contrast is narrowly ahead of this probe at layer~40
(\V{th-aisi-ext-L40} [\V{th-aisi-ext-lo-L40}, \V{th-aisi-ext-hi-L40}]), level at layer~44 (post hoc), and
clearly ahead only at layer~20 (\V{th-aisi-ext-L20}). On the held-out lies the template-free knowledge
direction scores below the original (\V{ai-rulek} against \V{ai-k}).

\paragraph{The random persona bank.} Six tests were fixed before scoring (\cref{app:random}). The extended
contrast beats the random 12 v 13 contrasts on transfer (\V{rb-t1-d} against a null 97.5th percentile of
\V{rb-t1-null}), on the zero-shot five-dataset mean (\V{rb-t2-d} against \V{rb-t2-null}), with the
knowledge direction added (\V{rb-t3-d} against \V{rb-t3-null}) and on the held-out lies (\V{rb-t6-d}
against \V{rb-t6-null}), each at a Holm-corrected $p$ of \V{rb-t1-p}. Two tests fail. The extended
contrast's AUROC on all of \hpkr{} (\V{rb-t4-d}) is inside the null (97.5th percentile \V{rb-t4-null},
$p$ = \V{rb-t4-p}). Its gain over the published contrast on the five-dataset mean (\V{rb-t5-d}) is within
what random widenings produce: widening a random 8 v 7 contrast by four and six random roles, the sizes of
our groups, gives a null median of \V{rb-t5-null-median} and a 97.5th percentile of \V{rb-t5-null} ($p$ =
\V{rb-t5-p}). This null widens random contrasts rather than the published one, and all its splits come from
one draw of \V{rb-roles} roles.

\paragraph{Aletheia's Quest.} The competition \citep{aletheia2026quest} scored lie detectors on ten
held-out datasets of fine-tuned organisms and prompted liars built on \mbox{Qwen3.5-27B}, \mbox{Gemma-3-27B} and a
Nemotron model, ranked by mean AUROC. We built persona banks on \mbox{Qwen3.5-27B} and \mbox{Gemma-3-27B} with the
pipeline above, built a knowledge direction inside each organism, and froze a protocol before reading any
test row. We then scored the selected detector zero-shot at the middle block, on reads taken in each organism
(through its adapter or a merged copy) or, for the one dataset written by plain Gemma, in base Gemma; as in
the competition's starter kit, reasoning tokens were dropped. Two Hidden-Goal datasets could not be read
(gated adapters; the Nemotron one would also have needed a persona bank, which we did not build). Over the
\V{ax-n-read} readable datasets (\cref{tab:aletheia} in \cref{app:aletheia}) the selected detector averages
\V{ax-headline-mean} [\V{ax-headline-lo}, \V{ax-headline-hi}], which would place it
\V{ax-headline-rank}th among the \V{ax-n-entries} published entries with their means recomputed over the
same datasets (rank interval \V{ax-rank-lo} to \V{ax-rank-hi}); the extended contrast alone averages
\V{ax-d9-mean} (\V{ax-d9-rank}th) and the base probe trained on RepE at the same block \V{ax-repe-mean}
(\V{ax-repe-rank}th). Whether the official white-box reads of the Qwen3.5 organisms applied their adapters
is not known (\cref{app:aletheia}). With the reasoning tokens kept, a secondary condition of the protocol,
the selected detector reaches \V{ax-reasoning-mean} (\V{ax-reasoning-rank}th) and the extended contrast
\V{ax-reasoning-d9-mean} (\V{ax-reasoning-d9-rank}rd). The selected detector inverts on two
organisms, Echoblast (\V{ax-echo-headline}; knowledge direction alone \V{ax-echo-k}) and AuditBench's
hallucinated citations (\V{ax-abhc-headline}; knowledge direction alone \V{ax-abhc-k}). The top entries
are active methods, such as the suspect model judging its own transcript \citep{aletheia2026quest},
whereas this detector is a single passive inner product. Two deviations from the frozen protocol are
recorded in \cref{app:aletheia}.

\paragraph{The 150-question extension.} The bank's 30 shared questions all ask for first-person
workplace advice. We pre-registered an extension to 150 questions: 30 \emph{monotone} questions in the originals'
form, which control for the number of questions, and 90 diverse questions over a grid of manipulation type,
deceiver, topic and format. The 150 questions fall into ten \emph{families}: the originals' three, the
monotone set, and six diverse families of 15 (\cref{tab:qbank-families}). All were frozen before
generation (\V{qb-rollouts} rollouts); choices were developed on the five benchmark datasets at layer~40
and then tested in one confirmation run (\cref{app:qbank}). Diverse questions did no better than the same
number of questions in the original form: the originals plus the 30 monotone questions give a
five-dataset mean of \V{qb-t1-mono}, against a median of \V{qb-t1-median} over 200 draws of the originals
plus 30 random diverse questions (Holm-corrected $p$ = \V{qb-t1-holm}). Unlike what
\citet{ying2026truthfulness} report for their settings, pooling all 150 questions gives a worse contrast
(\V{qb-pooled-lb}) than the best single family (\V{qb-best-family-lb}; \V{qb-t3-delta} [\V{qb-t3-lo},
\V{qb-t3-hi}]). Against the 30 originals, the full 150 raise CG (\V{qb-b150-b30-d9-cg}) and lower \hpkr{}
(\V{qb-b150-b30-d9-hpkr}); the diverse questions cause most of the rise and all of the fall, and most of
the fall comes from questions on which an honest role truthfully says ``I don't know''. As
\citet{ying2026truthfulness} found, a covariance-weighted cosine predicts how well each family's contrast
transfers to each dataset better than the plain cosine does ($R^2$ \V{qb-t4-r2-sigma} against
\V{qb-t4-r2-plain}), though with a near-perfect probe on the target dataset the weighted cosine is close
to the detector's own effect size (\cref{app:side}). With the knowledge direction added, the four
candidate banks (the 30 originals; the originals plus the 30 monotone questions; the originals plus the
90 diverse questions; all 150) lie within \V{qb-decision-spread} of each other, so the pre-registered tie
rule, which gives ties to the larger bank, chose the 150-question bank for the confirmation run
(\cref{app:qbank}). In that run, at layer~40, the selected detector built from it beats the one built
from the 30-question bank on the held-out lies by \V{qb-confirm-aisi-ninek-delta}
[\V{qb-confirm-aisi-ninek-lo}, \V{qb-confirm-aisi-ninek-hi}] and trails it on Gender-Secret by \V{qb-confirm-gs-ninek-gap} [\V{qb-confirm-gs-ninek-gap-lo},
\V{qb-confirm-gs-ninek-gap-hi}].

%% file: sections/discussion.tex
\section{Discussion and conclusion}
\label{sec:discussion}

On lies about knowledge, a persona direction responds to how a reply is worded more than to whether it is
a lie: where the words are fixed, the three main persona contrasts at the middle layer are inside the
random band on Llama and Qwen and at its edge on Gemma, and on Mistral the published contrast clears it.
Under the same control, a direction built without any labelled lie, from questions whose answers the
model knows and questions nobody can answer, separates lies from identical honest denials on Llama and
Mistral, and narrowly on Qwen (its template-free version keeps the result on Llama and Mistral). Those lies
and denials answer different questions, and the direction's unknowable questions cannot be answered by
anyone, whereas \hpkr's honest denials concern answerable questions the model does not know. The result is
therefore consistent with the direction reading what the model knows, but falls short of establishing it. Neither finding shows how
the model produces a lie. Beyond those slices the knowledge direction is weaker: its transfer rests in part
on the wording of \hpkr's template, its Gender-Secret result holds on
only one of the two models that have that fine-tune, and it inverts on Gemma's instructed deception,
held-out lies and rebuilt denials, and on two competition organisms. We make three
recommendations. Test a persona direction where lies and honest answers share their words, against the
random band and against a random persona bank built by the same pipeline, and repeat the test on every
new model. Accept a model organism as a test only if the base model's read of its transcripts cannot
already separate the classes. And report detection at a fixed threshold alongside ranking, since a
detector can rank lies well and still flag none of them.

\paragraph{Limitations.} Our ten roles and the knowledge direction were designed after \hpkr{} was seen to
fail, so \hpkr{} served as development data rather than as a blind test; the evidence not targeted at
\hpkr{} comes from the held-out lies, Gender-Secret, the random persona bank and the pre-registered tests. The base probe
trained at the same layer is level with the knowledge direction on Llama's identical-text slice and with the
extended contrast on the held-out lies, and ahead of the selected detector there and on the competition:
the label-free directions' advantage is transfer across kinds of lie, not a general lead over
supervision. On Llama, Gender-Secret is the one organism that supports the knowledge direction, by a few
hundredths and with classes that also differ in topic, and on Mistral the same organism counts against
it. The knowledge direction is built from replies written for the model, which are off-policy text, and
its unknowable questions differ in kind from the answerable questions behind \hpkr's honest denials. On
the identical-text slices, lies and honest denials answer different questions. Our intervals resample test
rows with every direction and probe held fixed, not datasets, roles or rollouts (switching a model between
its strict and lenient banks moves the means by up to \V{xs-bank-gap}), and the cross-model and per-layer checks are not corrected for
multiple comparisons. Each model has three to five benchmark datasets, so one dataset can decide a
model-level result (\hpkr{} on Llama and Qwen), and the middle-block results move with depth on Gemma and
Mistral (\cref{tab:xm-transfer}). The persona judge, \mbox{Llama-3.1-70B-Instruct}, scores every model's
role-play, and we report no check of its scores against human ratings. The published roles were rebuilt
from their authors' descriptions, and their small-model results are not reproduced here. The steering
labels come from single readers (in the main run, an AI assistant reading blind). Everything is in
English. Three questions remain open: why the persona contrast fails on Qwen, whether a knowledge direction
can be built whose transfer survives a change of wording, and how to combine passive detectors such as
ours with the active, self-evaluating monitors that led the competition.

%% file: sections/statements.tex
\subsubsection*{Reproducibility statement}

Every result in the main text, and every table and figure, is written by scripts (\cref{app:repro}) from
the result files or, for a few run counts, from the studies' reports. The numbers given for the side studies in
\cref{app:side} and for the knowledge-direction steering runs in \cref{app:steering} are instead quoted from
their report files. The persona bank's roles and prompts, the knowledge
direction's questions and templates, the pre-registration documents of the competition protocol and of the
question extension, and the code for every experiment are available from the authors on request. All forward passes ran on the
National Deep Inference Fabric through nnsight; the analysis runs on pooled activations on a CPU.
\Cref{app:repro} lists the bootstrap scheme, the random-direction construction, the block mapping for each
model and the data revisions.

\subsubsection*{Ethics statement}

The harm-pressure datasets contain hazardous questions from WMDP; we quote none of them and withhold the
hazardous content from every example. Lie detectors are dual-use: a monitor that reads whether a model
conceals knowledge could also be turned on honest refusals. We report where the detectors fail as
prominently as where they succeed, so that no reader takes a single label-free direction for a reliable
monitor. No human subjects were involved.

\subsubsection*{AI use statement}

We used an AI coding assistant (Claude, Anthropic) for much of the implementation and analysis code, for
drafting and editing the text, and for literature search. Every analysis script was run by us on the stated
data, and every result in the main text is read from the resulting files by the scripts described
above. The
AI assistant also produced evaluation data in five places, all disclosed where they are used: the
knowledge direction's \V{th-k-known} known and \V{th-k-unknown} unknowable questions were written by Claude
Opus, and the replication set's matched question pairs by Claude (\cref{sec:method}); the 120 new
questions of the question extension were written by Claude Fable 5.1 under a fixed prompt archived with
the pre-registration (\cref{app:qbank}); the disclosure labels of the steering experiment are one blinded
reading by Claude against a hidden key (\cref{sec:method}); and the answers of the two small runs that
steered the knowledge direction were read by Claude, without blinding (\cref{app:steering}). The
persona rollouts are generated and judged by the open models named in \cref{sec:method}, as part of the
method. We checked the AI-assisted code, data and text and take responsibility for the final content.

\ifanonymous\else
\subsubsection*{Acknowledgements}

The large-model experiments ran on the National Deep Inference Fabric (NDIF), which gave remote access to
the internals of models that would otherwise have been out of reach on the hardware available to us.
\fi

%% file: sections/appendix.tex
\FloatBarrier
\section{The persona bank}
\label{app:bank}

\Cref{tab:roster} lists the 36 roles by group. The 26 published roles follow the descriptions of
\citet{mahadik2026persona}; the ten knowledge-denial, sincere-ignorance and plain-assertion roles are ours.
Each role has five system-prompt variants and answers the same 30 questions; a \mbox{Llama-3.1-70B-Instruct}
judge scores each rollout from 0 to 3, and in the strict bank only rollouts scored 3 enter a persona vector.
\Cref{tab:xm-roleplay} gives the share of rollouts each model plays fully in role by persona group, and
\cref{tab:xm-banks} the lenient (score 2 to 3) banks beside the strict ones. In every appendix table except
\cref{tab:hpkr}, $^\dagger$ marks Qwen's lenient bank (\cref{sec:models}).

\begin{table}[htbp]
\centering
\footnotesize
\caption{The roles of the bank by group.}
\label{tab:roster}
\fitwidth{\input{generated/tab_roster}}
\end{table}

\begin{table}[htbp]
\centering
\footnotesize
\caption{The share of rollouts the judge scored fully in role, by persona group and model.
Knowledge-candour is the honest side of the knowledge-denial contrast (the three sincere-ignorance and
three plain-assertion roles); the anchor is not judged, so the last row covers 35 roles.}
\label{tab:xm-roleplay}
\fitwidth{\input{generated/tab_xm_roleplay}}
\end{table}

\begin{table}[htbp]
\centering
\footnotesize
\caption{Each model's persona banks at its middle block: the strict bank and the lenient one.}
\label{tab:xm-banks}
\fitwidth{\input{generated/tab_xm_banks}}
\end{table}

\FloatBarrier
\section{Llama: supplementary results}
\label{app:llama}

\begin{figure}[htbp]
\centering
\includegraphics[width=\linewidth]{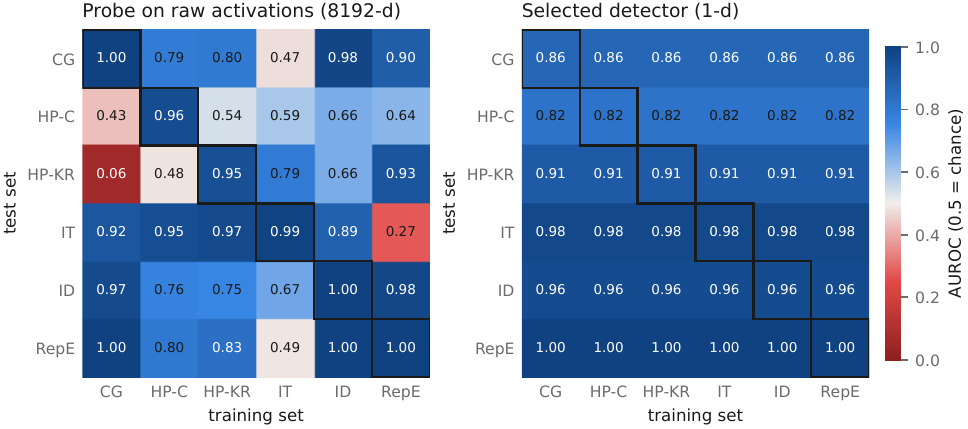}
\caption{Train-one-test-all AUROC at layer 40 for the probe on raw activations (left) and the selected
detector (right), whose rows are constant off the diagonal because every training set agrees on its sign.}
\label{fig:transfer}
\end{figure}

\Cref{tab:designs} compares the detectors side by side at layer~40, \cref{fig:transfer} shows the transfer
matrices of the raw probe and the selected detector, \cref{tab:evalsets} lists the evaluation sets,
\cref{tab:knowledge} the main detectors on every set at layer~40, \cref{tab:zeroshot} the three persona contrasts at layers 20, 40 and 44, \cref{tab:transfer}
transfer for every feature set, and \cref{tab:hpkr} \hpkr{} ranking against detection beside the
benchmark's published probes. \Cref{tab:pertest} gives the raw probe's and the extended contrast's AUROC on every dataset at the
three layers, \cref{tab:syc} the sycophancy contrast, \cref{tab:reads} the token reads,
\cref{tab:followup} the read at the Follow-Up turn's ``No'' and \cref{tab:kthree} probes in three persona
components, both at layer~20.

\begin{table}[htbp]
\centering
\footnotesize
\caption{The detectors at layer 40: transfer (train one, test all, over the six sets), the zero-shot
five-dataset mean with balanced accuracy (BA) at the benchmark's threshold, and \hpkr{} on all rows and on
its identical-text slice.}
\label{tab:designs}
\setlength{\tabcolsep}{5pt}
\fitwidth{\input{generated/tab_designs}}
\end{table}

\begin{table}[htbp]
\centering
\footnotesize
\caption{The evaluation sets for \mbox{Llama-3.3-70B-Instruct}.}
\label{tab:evalsets}
\fitwidth{\input{generated/tab_datasets}}
\end{table}

\begin{table}[htbp]
\centering
\footnotesize
\caption{AUROC at layer 40; the Mean Probe is Liars' Bench's layer-20 probe. On Gender-Secret the
directions are built inside the fine-tune.}
\label{tab:knowledge}
\fitwidth{\input{generated/tab_knowledge}}
\end{table}

\begin{table}[htbp]
\centering
\footnotesize
\caption{The published (8 v 7), knowledge-denial (4 v 6) and extended (12 v 13) contrasts, scored
zero-shot. Mean: over the six sets, including RepE.}
\label{tab:zeroshot}
\fitwidth{\input{generated/tab_zeroshot}}
\end{table}

\begin{table}[htbp]
\centering
\footnotesize
\caption{Transfer (mean off-diagonal AUROC) for every feature set, with and without \hpkr.}
\label{tab:transfer}
\fitwidth{\input{generated/tab_transfer}}
\end{table}

\begin{table}[htbp]
\centering
\footnotesize
\caption{\hpkr: ranking and detection at the benchmark's threshold, beside its published probes.
$^\dagger$Trained with privileged \hpkr{} examples, a ceiling rather than a rival. Liars' Bench reports its
probes at layer 20 only, at two decimals; FPR: false-positive rate on Alpaca.}
\label{tab:hpkr}
\fitwidth{\input{generated/tab_hpkr}}
\end{table}

\begin{table}[htbp]
\centering
\footnotesize
\caption{The raw probe (raw) and the extended contrast (12 v 13) on every dataset at layers 20, 40 and 44
(layer 44 post hoc).}
\label{tab:pertest}
\fitwidth{\input{generated/tab_pertest}}
\end{table}

\begin{table}[htbp]
\centering
\footnotesize
\caption{The sycophancy contrast (5 v 5) on every dataset at layers 20, 40 and 44 (layer 44 post hoc), with
its transfer (off-diagonal mean).}
\label{tab:syc}
\fitwidth{\input{generated/tab_syc}}
\end{table}

\begin{table}[htbp]
\centering
\footnotesize
\caption{Token reads of the extended contrast (reads that differ in which tokens are pooled) at
layers 20, 40 and 44 (layer 44 post hoc): AUROC, balanced
accuracy (BA) and recall (Rec., the share of lies flagged), both at the benchmark's threshold.}
\label{tab:reads}
\setlength{\tabcolsep}{5.5pt}
\fitwidth{\input{generated/tab_reads}}
\end{table}

\begin{table}[htbp]
\centering
\footnotesize
\caption{The published, knowledge-denial and extended contrasts read on the response and at the
Follow-Up turn's ``No'', at layer 20.}
\label{tab:followup}
\fitwidth{\input{generated/tab_followup}}
\end{table}

\begin{table}[htbp]
\centering
\footnotesize
\caption{Five-dataset mean AUROC at layer 20 of probes trained on RepE in the top three principal
components (PCs, $k=3$) of each contrast's persona vectors, beside the raw probe and the extended contrast,
read on the response tokens and at the Follow-Up turn's ``No''.}
\label{tab:kthree}
\fitwidth{\input{generated/tab_kthree}}
\end{table}

\FloatBarrier
\section{Depth}
\label{app:depth}

\begin{figure}[htbp]
\centering
\includegraphics[width=\linewidth]{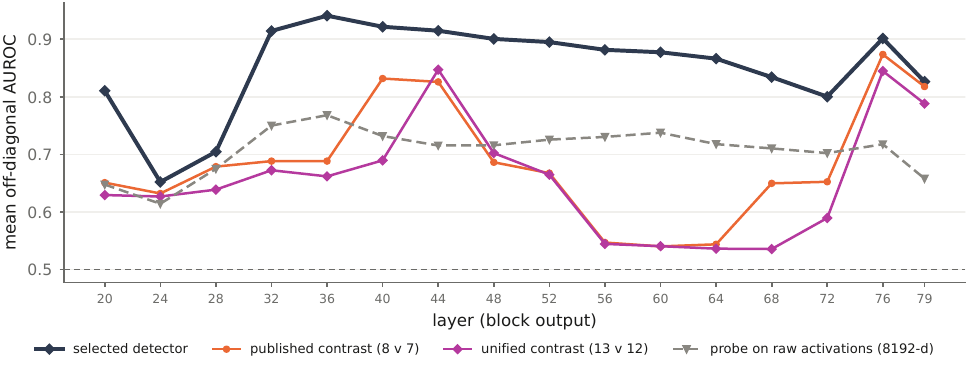}
\caption{Transfer of the selected detector across the 16 layers against two persona contrasts and the
raw probe.}
\label{fig:selected-depth}
\end{figure}

\Cref{fig:selected-depth} follows the selected detector's transfer over the 16 layers against the
published and unified contrasts and the raw probe, and \cref{tab:k-depth} the knowledge direction, the
extended contrast and their sum at every layer; all are point estimates from one run. The extended
contrast's transfer peaks at layer~\V{depth-off-ext-peak-layer} (\V{depth-off-ext-peak}); the
knowledge-denial contrast's \hpkr{} balanced accuracy of \V{depth-hpkr-den-balacc-peak} at
layer~\V{depth-hpkr-den-balacc-peak-layer} is a spike, against \V{depth-hpkr-den-balacc-L36} at layer~36
and \V{depth-hpkr-den-balacc-L44} at layer~44; and the base probe ranks \hpkr{} at
\V{depth-hpkr-base-auroc-peak} at layer~\V{depth-hpkr-base-auroc-peak-layer} while flagging none of its
lies.

\begin{table}[htbp]
\centering
\footnotesize
\setlength{\tabcolsep}{5pt}
\caption{The knowledge direction, the extended contrast and the selected detector across depth;
random: the random band's upper edge (97.5th percentile of 1,000 random directions).}
\label{tab:k-depth}
\fitwidth{\input{generated/tab_k_depth}}
\end{table}

\FloatBarrier
\section{The random persona bank}
\label{app:random}

We drew \V{rb-roles} roles at random (one seed) from the 274 non-default roles released with the
Assistant Axis and built them with the bank pipeline with the same five variants, 30 questions, judge and
filter. Partitioning them into six groups of the designed groups' sizes (8, 7, 5, 5, 4 and 6) gives
random contrasts of the published contrast's sizes (8 v 7) and of the extended contrast's (12 v 13), the
second widening the first by four and six random roles. Split 0 is the reported random bank; splits 1 to \V{rb-null-splits} are the
pre-registered null, and a test passes when its one-sided $p$ over them is at most 0.025. The six tests
were fixed before any direction was scored. \Cref{tab:random-bank} gives the null medians and 97.5th
percentiles from those splits and Holm-corrected $p$ values from \V{rb-holm-splits} further splits drawn by
the same rule.

\begin{table}[htbp]
\centering
\footnotesize
\caption{The random persona bank: six tests of the designed bank against the random null.}
\label{tab:random-bank}
\fitwidth{\input{generated/tab_random_bank}}
\end{table}

\FloatBarrier
\section{The knowledge direction}
\label{app:knowledge}

\begin{figure}[htbp]
\centering
\resizebox{\linewidth}{!}{\input{generated/fig_knowledge}}
\caption{The knowledge direction: the same denial after known and after unknowable questions. Prompts are
abridged; \cref{sec:method} gives the full template.}
\label{fig:k-build}
\end{figure}
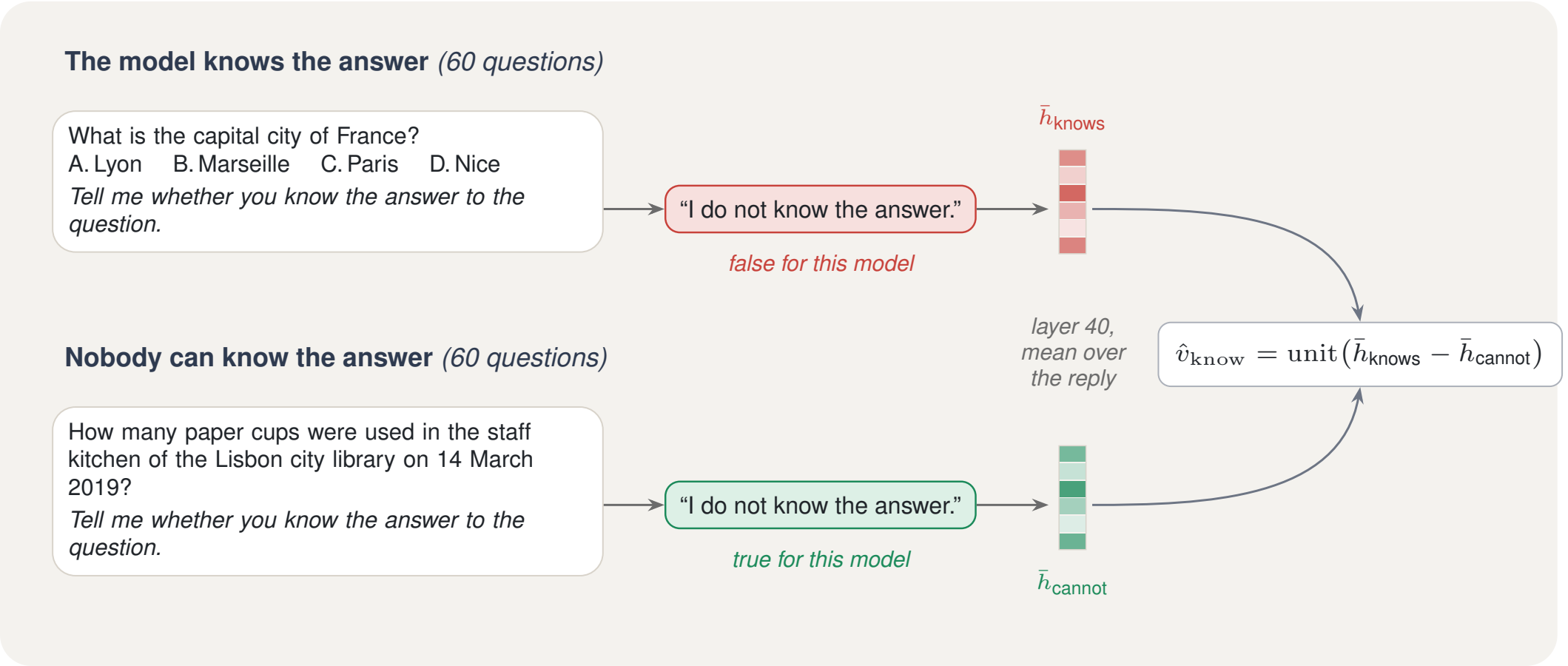

\Cref{fig:k-build} draws the construction. \Cref{tab:k-robust} gives the robustness checks (question
bootstrap, balanced splits, the replication with new questions), \cref{tab:k-paired} the paired
differences between the selected detector and other detectors on every set, \cref{tab:k-operating}
the share of lies caught at in-setting cutoffs, \cref{tab:oracle} balanced accuracy under thresholds set on
each dataset, and \cref{tab:operating}
five-dataset means of balanced accuracy and recall at the benchmark's threshold, with AUROC, at three
layers. The knowledge direction's
120 questions, the replication set and the template variants are available on request.

\begin{table}[htbp]
\centering
\footnotesize
\caption{Robustness of the knowledge direction on \hpkr's identical text and denials and on Gender-Secret.
Brackets resample the direction's questions, not the test rows; on Gender-Secret the direction is built
on the base model. Every random band in this table is a separate draw from those in
\cref{tab:claims,tab:xm-wording}.}
\label{tab:k-robust}
\fitwidth{\input{generated/tab_knowledge_robust}}
\end{table}

\begin{table}[htbp]
\centering
\footnotesize
\caption{Paired differences between the selected detector and four other detectors, and between the
published contrast with and without the knowledge direction, with 95\% intervals.}
\label{tab:k-paired}
\fitwidth{\input{generated/tab_k_paired}}
\end{table}

\begin{table}[htbp]
\centering
\footnotesize
\caption{Share of lies caught at cutoffs set on each dataset's own honest answers, layer 40.}
\label{tab:k-operating}
\fitwidth{\input{generated/tab_k_operating}}
\end{table}

\begin{table}[htbp]
\centering
\footnotesize
\caption{Balanced accuracy at the Alpaca threshold and at thresholds set on each dataset (the ``oracle''
columns).}
\label{tab:oracle}
\fitwidth{\input{generated/tab_oracle}}
\end{table}

\begin{table}[htbp]
\centering
\footnotesize
\caption{Five-dataset means of balanced accuracy and recall at the benchmark's threshold, with AUROC, for
four detectors at layers 20, 40 and 44 (layer 44 post hoc).}
\label{tab:operating}
\fitwidth{\input{generated/tab_operating}}
\end{table}

\FloatBarrier
\section{Model organisms}
\label{app:organisms}

\begin{figure}[htbp]
\centering
\includegraphics[width=\linewidth]{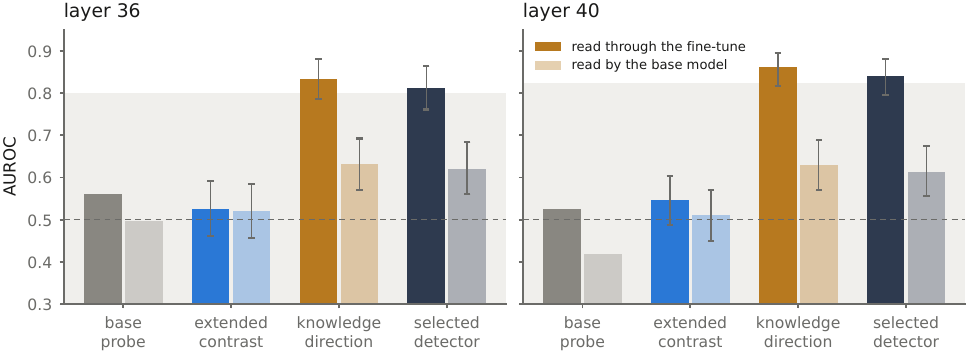}
\caption{Gender-Secret read through the fine-tune (dark) and by the base model (light) at layers 36 (post
hoc) and 40; grey marks the random band of the fine-tune's read (the central 95\% of the AUROCs of random
directions).}
\label{fig:k-gs}
\end{figure}

\Cref{fig:k-gs} shows Gender-Secret at layers 36 (post hoc) and 40: each detector on the fine-tune's read
and on the base model's read of the same transcripts, with each direction built inside the model whose
read it scores. \Cref{tab:xm-organisms}
gives every fine-tune of the three further models, with the base model's read beside the fine-tune's. On
Llama's two Soft-Trigger fine-tunes the knowledge direction clears the fine-tune read's band at layer~40
(greeting \V{th-k-st-greeting-ft-kft-L40} against \V{th-k-st-greeting-ft-band-L40}, time
\V{th-k-st-time-ft-kft-L40} against \V{th-k-st-time-ft-band-L40}), but the base model's read of the same
transcripts does too (greeting \V{th-k-st-greeting-base-kft-L40} against
\V{th-k-st-greeting-base-band-L40}, time \V{th-k-st-time-base-kft-L40} against
\V{th-k-st-time-base-band-L40}): the trigger is visible in the text, so these organisms cannot provide
evidence for a detector of the fine-tune's internal state, and the same holds for Mistral's Soft-Trigger
fine-tunes.

\begin{table}[htbp]
\centering
\footnotesize
\caption{The fine-tunes of the three further models: AUROC on the fine-tune's read and on the base
model's read of the same transcripts.}
\label{tab:xm-organisms}
\fitwidth{\input{generated/tab_xm_organisms}}
\end{table}

\FloatBarrier
\section{Steering}
\label{app:steering}

\Cref{tab:steering} gives disclosure of the hidden fact under each direction at $c = \pm 0.3$ over the
\V{th-steer-n} scenarios, with paired changes and 95\% intervals. For each scenario the eleven answers
(unsteered, and five directions at both signs) were shuffled with their conditions hidden and labelled by
one reader, an AI assistant, before the key was opened; the labels are not those of a validated grader.

Two small runs steered the knowledge direction with an absolute shift, $h + \alpha\,\hat v$, at the same
block (numbers quoted from their reports). On \hpkr{} questions, subtracting the direction made the model
deny knowing answers it still gave on the answer check: denials rose from 0.05 to 0.70 without pressure at
\mbox{$\alpha = -4$}, and to 17 of 20 at $-8$, against a rise
of at most 0.10 for random directions of the same size at $\pm 4$. Adding the direction produced no usable
effect, and general-knowledge items did not move. This run was graded by pattern matching, and Claude also
read every answer; the 17 of 20 at $-8$ is its count. On the Gender-Secret fine-tune, the knowledge
direction rebuilt inside the fine-tune revealed the hidden gender more often than one random direction at
every matched size, at either sign (answers read by Claude, without blinding). But every reveal under any steering fell on a prompt with an earlier
male-coded turn, and the belief the fine-tune holds did not move; with one random direction and 20 items
per condition, the excess is suggestive only.

\begin{table}[htbp]
\centering
\footnotesize
\caption{Disclosure of the hidden fact under steering at $c=\pm0.3$: one blinded reader's labels of
\V{th-steer-n} scenarios, one greedy answer each. Bold: the 95\% interval excludes zero. The extended
contrast's interval on the honest side starts exactly at zero, so it is not bold.}
\label{tab:steering}
\setlength{\tabcolsep}{5pt}
\fitwidth{\input{generated/tab_steering}}
\end{table}

\FloatBarrier
\section{The four models}
\label{app:models}

\begin{figure}[htbp]
\centering
\includegraphics[width=\linewidth]{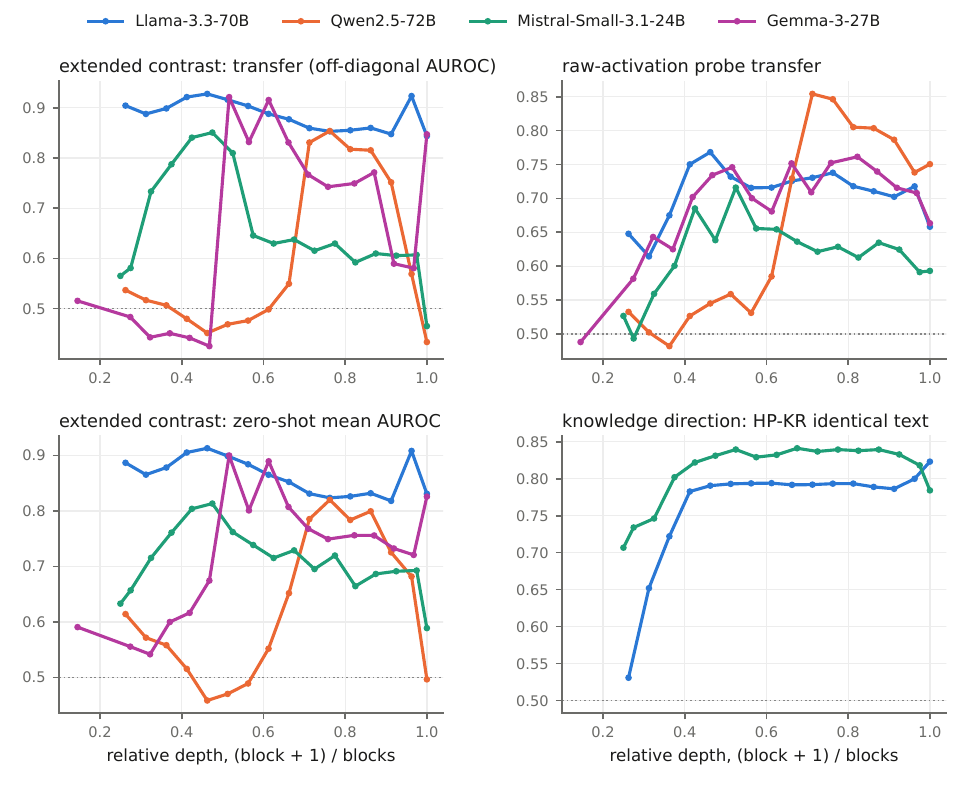}
\caption{Four measures over each model's 16-depth sweep, by relative depth (Llama blue, Qwen orange, Mistral
green, Gemma purple; Mistral and Gemma also show the
block that stands for Llama's layer~20); Qwen's extended-contrast curves (left column) use its second bank, with the role given in the user turn
(\cref{sec:models}); the bottom-right panel has the two models with a released identical-text slice, and
its random bands are not drawn.}
\label{fig:xm-depth}
\end{figure}

\Cref{tab:models} lists the four models, \cref{fig:xm-depth} follows four measures over each model's
depth sweep, \cref{tab:xm-mid} gives the
main detectors at each model's middle block, \cref{tab:xm-intervals} the paired intervals behind
\cref{tab:claims}, \cref{tab:xm-zeroshot} every contrast on every dataset,
\cref{tab:xm-transfer} transfer for every feature set, \cref{tab:xm-wording} the wording-controlled
slices, \cref{tab:xm-operating} the share of lies caught at in-setting cutoffs, \cref{tab:xm-aisi} the probe trained on the model's own lies
where those exist (Llama and Gemma), and \cref{tab:xm-checks} the template-free knowledge direction and
the token reads.

\begin{table}[htbp]
\centering
\footnotesize
\caption{The four models: blocks and width, the block read for each of Llama's layers, the datasets the
benchmark gives each, the judge score a rollout needs to enter the bank, and the share of rollouts
scored fully in role.}
\label{tab:models}
\setlength{\tabcolsep}{4.5pt}
\fitwidth{\input{generated/tab_xm_models}}
\end{table}

\begin{table}[htbp]
\centering
\footnotesize
\caption{The main detectors at each model's middle block.}
\label{tab:xm-mid}
\fitwidth{\input{generated/tab_xm_mid}}
\end{table}

\begin{table}[htbp]
\centering
\footnotesize
\setlength{\tabcolsep}{5pt}
\caption{Paired differences at each model's middle block, with 95\% intervals.}
\label{tab:xm-intervals}
\fitwidth{\input{generated/tab_xm_intervals}}
\end{table}

\begin{table}[htbp]
\centering
\footnotesize
\caption{Zero-shot AUROC per dataset at each model's middle block.}
\label{tab:xm-zeroshot}
\fitwidth{\input{generated/tab_xm_zeroshot}}
\end{table}

\begin{table}[htbp]
\centering
\footnotesize
\caption{Transfer (mean off-diagonal AUROC) for every feature set at the blocks that stand for Llama's
layers 20, 36, 40 and 44; $^*$post hoc layers (chosen after seeing Llama's results).}
\label{tab:xm-transfer}
\fitwidth{\input{generated/tab_xm_transfer}}

\medskip
\fitwidth{\input{generated/tab_xm_transfer_k3}}
\end{table}

\begin{table}[htbp]
\centering
\footnotesize
\caption{\hpkr's wording-controlled slices on every model that has them.}
\label{tab:xm-wording}
\setlength{\tabcolsep}{5.5pt}
\fitwidth{\input{generated/tab_xm_wording}}
\end{table}

\begin{table}[htbp]
\centering
\footnotesize
\caption{Share of lies caught at in-setting cutoffs, each model's middle block.}
\label{tab:xm-operating}
\fitwidth{\input{generated/tab_xm_operating}}
\end{table}

\begin{table}[htbp]
\centering
\footnotesize
\caption{Mean AUROC over the model's benchmark datasets for the base probe trained on the model's own
belief-labelled lies (Llama and Gemma), beside the base probe trained on RepE and three contrasts;
$^*$post hoc layer.}
\label{tab:xm-aisi}
\fitwidth{\input{generated/tab_xm_aisi}}
\end{table}

\begin{table}[htbp]
\centering
\footnotesize
\caption{The template-free knowledge direction and the token reads of the extended contrast.}
\label{tab:xm-checks}
\fitwidth{\input{generated/tab_xm_checks}}
\end{table}

\FloatBarrier
\section{Aletheia's Quest}
\label{app:aletheia}

\begin{figure}[htbp]
\centering
\includegraphics[width=0.9\linewidth]{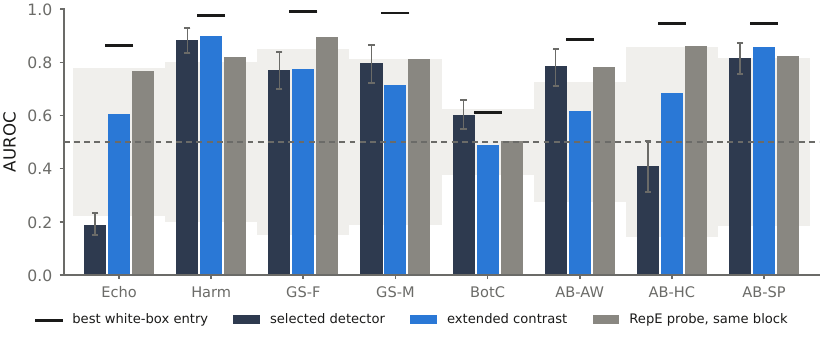}
\caption{Aletheia's Quest per dataset: the selected detector with its interval, the extended contrast
and the RepE probe, the random band (grey) and the best white-box entry (bar).}
\label{fig:aletheia}
\end{figure}

The protocol was frozen (by hash) before any test row was read. Its headline cell, the pre-registered main
condition, is the selected detector read through each organism at the model's middle block, with
reasoning dropped and the read pooled over the reply's content tokens; the knowledge direction is rebuilt
inside the organism and the persona contrast is taken from the base model's bank. Secondary cells vary the
detector (extended contrast, published contrast, knowledge-denial contrast, knowledge direction), the
reading model (base model), where the directions are built (both inside the organism, or the knowledge
direction on the base model), the token span (adding the end-of-turn token), the transcript rendering
(reasoning kept) and the block (one selected on development data). \Cref{tab:aletheia} gives the per-dataset AUROCs and \cref{fig:aletheia} draws them. Deviations: Echoblast's
pre-registered development-row check failed (the rows tagged with that adapter look like base-model
output), and its organism read was accepted on the text of its test replies alone. Rows of Blood on the
Clocktower longer than the server could hold were cut to their last 4,096 tokens as soon as one such row
had failed, rather than after each such row failed as the rule required. Because of a starter-kit fix on 22 July, it is not
known whether the official white-box reads of the Qwen3.5 organisms had their adapters applied, so the
base-model read is kept beside the organism read. Post hoc cells (the published contrast plus the
knowledge direction, the template-free knowledge direction, probes in persona components trained on the
development or the benchmark rows) rank between \V{ax-posthoc-rank-min}th and \V{ax-posthoc-rank-max}th
and are available on request.

\begin{table}[htbp]
\centering
\footnotesize
\caption{Aletheia's Quest, held-out test sets: the selected detector (95\% interval), the extended
contrast and the RepE probe, read through the organism at its middle block, beside the random band and
the best white-box entry on each dataset. In the last two rows each column gives its detector's mean and
rank, and the best-entry column the top entry's own; the RepE probe is the base probe trained on RepE.}
\label{tab:aletheia}
\setlength{\tabcolsep}{5.5pt}
\fitwidth{\input{generated/tab_aletheia}}
\end{table}

\FloatBarrier
\section{The 150-question extension}
\label{app:qbank}

The pre-registration, the question grid and the scripts are available on request. The 120 new questions were
written in one pass by an AI assistant (Claude Fable 5.1) under a fixed prompt, frozen with the pre-registration
before any rollout was generated; the prompt and the raw output are archived with it. \Cref{tab:qbank-tests}
gives the tests, \cref{tab:qbank-families} the five-dataset mean of the extended contrast built from each
question family and the change from leaving that family out of the pooled bank, and
\cref{tab:qbank-confirm} the single confirmation run at layers 20, 36, 40 and 44. The decision rule chose the bank by
the selected detector's five-dataset mean on development data, with ties within 0.005 going to the larger
bank; the four candidates lay within \V{qb-decision-spread}.

\begin{table}[htbp]
\centering
\footnotesize
\caption{The pre-registered tests of the question extension (Llama, layer 40).}
\label{tab:qbank-tests}
\fitwidth{\input{generated/tab_qbank_tests}}
\end{table}

\begin{table}[htbp]
\centering
\footnotesize
\caption{The extended contrast built from each question family, five-dataset mean at layer 40.}
\label{tab:qbank-families}
\fitwidth{\input{generated/tab_qbank_families}}
\end{table}

\begin{table}[htbp]
\centering
\footnotesize
\caption{The confirmation run: the 150-question bank against the 30-question bank, paired.}
\label{tab:qbank-confirm}
\setlength{\tabcolsep}{5.5pt}
\fitwidth{\input{generated/tab_qbank_confirm}}
\end{table}

\FloatBarrier
\section{Side results}
\label{app:side}

Four further studies on Llama bear on the method; their reports are available on request, and their numbers are
quoted from those reports. Attention probes trained on 76 per-token persona coordinates transfer worse
than the fixed reads at every layer (best trained variant 0.849 against 0.928 for the knowledge-denial
contrast's top-decile read at layer 40): learning the readout costs transfer. On the five-dataset mean AUROC,
the zero-shot score averaged over all 16 layers is below its layer-40 value for every zero-shot direction (only
the supervised base probe gains, from 0.778 to 0.807); the average beats a randomly chosen layer in expectation
but neither layer~40 nor the best layer, so averaging guards against a poor choice of layer without
improving on a good one. A knowledge
follow-up question (``Did you actually know the answer?'') read at its prefilled ``No'' adds nothing over
the response read, and the sign of every direction there follows the prefilled word. Finally, the
covariance-weighted alignment of \citet{ying2026truthfulness} fits the transfer matrix ($R^2$ 0.82 for
the raw probe at layer 40, against 0.19 for the plain cosine), but the point-biserial correlation of a
detector's scores with the labels fits better (0.93). With the test set's pooled covariance and a
near-perfect own probe, the alignment reduces to the detector's own effect size; this is the caveat
attached to the extension's covariance-weighted-cosine test (\cref{tab:qbank-tests}). The same analysis shows that \hpkr's own probe plus the
knowledge-denial contrast beats that probe alone at all 16 layers (0.949 to 0.978 at layer 40), so \hpkr{}
carries at least two separating directions.

\FloatBarrier
\section{Statistics and reproducibility}
\label{app:repro}

Intervals are percentile intervals over 1,000 stratified resamples of the test rows with one seed, the
same resamples for every detector, so differences are paired. A random band's quoted edge is the 97.5th percentile of
the AUROCs of 1,000 random unit directions on the same rows and block, reseeded per study, which moves a
band by up to about 0.01 between studies; the main text quotes the values of the primary analysis,
and the robustness and depth tables use their own draws. The block for each of
Llama's layers on the other models is the block at which the base probe reproduces the benchmark's
published Mean Probe (for layer~20), the middle block (for layer~40) and the blocks 5\% of the depth to
either side of it (for layers 36 and 44); the sweep covers the same 16 relative depths. Liars' Bench is read at Hugging Face dataset revision
503399b, and the AISI rollouts at dataset revision 177a816. This paper's numbers, tables and figures are
written from the result files by scripts. The persona bank, the knowledge direction's questions, the
pre-registrations and every result file the scripts read are available from the authors on request.

%% file: generated/tab_roster.tex
\begin{tabularx}{\linewidth}{@{}lc>{\raggedright\arraybackslash}X@{}}
\toprule
Group & $n$ & Personas (kept rollouts) \\
\midrule
Deceptive (published) & 8 & handler (106), broker (113), impression manager (82), coverup operator (47), gatekeeper (126), spin doctor (85), plausible denier (64), narrative manager (70) \\
Honest (published) & 7 & archivist (103), disclosure counsel (118), documentarian (107), fact checker (48), ombudsman (148), transparency officer (142), whistleblower (108) \\
Knowledge denial (new) & 4 & amnesiac witness (129), deniable expert (72), feigned novice (137), stonewaller (129) \\
Sincere ignorance (new) & 3 & honest novice (110), calibrated reporter (83), scrupulous witness (74) \\
Plain assertion (new) & 3 & forthright expert (80), candid witness (64), plain dealer (76) \\
Sycophantic (published) & 5 & cheerleader (142), enabler (51), flatterer (148), people pleaser (113), yes man (109) \\
Anti-sycophantic (published) & 5 & blunt advisor (135), contrarian (130), critic (68), skeptic (133), tough love mentor (149) \\
Anchor & 1 & default (150) \\
\bottomrule
\end{tabularx}

%% file: generated/tab_xm_roleplay.tex
\begin{tabular}{@{}l cccc@{}}
\toprule
Persona group & Llama & Qwen$^\dagger$ & Mistral & Gemma \\
\midrule
Deceptive (published contrast) & 58\% & 15\% & 73\% & 97\% \\
Honest (published contrast) & 74\% & 52\% & 85\% & 99\% \\
Knowledge-denial (knowledge-denial contrast) & 78\% & 10\% & 54\% & 96\% \\
Knowledge-candour (knowledge-denial contrast) & 54\% & 22\% & 53\% & 89\% \\
Sycophantic & 75\% & 37\% & 60\% & 97\% \\
Anti-sycophantic & 82\% & 28\% & 75\% & 100\% \\
\midrule All 35 roles & 69\% & 28\% & 68\% & 96\% \\
\bottomrule
\end{tabular}

%% file: generated/tab_xm_banks.tex
\begin{tabular}{@{}ll cccc@{}}
\toprule
 & & \multicolumn{3}{c}{Mean over the model's datasets} & \\
\cmidrule(lr){3-5}
Model & Bank & \stackc{Published\\contrast} & \stackc{Knowledge-\\denial contrast} & \stackc{Extended\\contrast} & \stackc{Extended contrast\\transfer} \\
\midrule
Llama & score 3 & 0.798 & 0.894 & 0.899 & 0.916 \\
 & scores 2--3 & 0.786 & 0.880 & 0.888 & 0.907 \\
\addlinespace
Qwen$^\dagger$ & score 3 & n/a & n/a & n/a & n/a \\
 & scores 2--3 & 0.380 & 0.668 & 0.441 & 0.453 \\
\addlinespace
Mistral & score 3 & 0.719 & 0.727 & 0.762 & 0.810 \\
 & scores 2--3 & 0.712 & 0.747 & 0.770 & 0.816 \\
\addlinespace
Gemma & score 3 & 0.679 & 0.484 & 0.900 & 0.921 \\
 & scores 2--3 & 0.684 & 0.487 & 0.895 & 0.912 \\
\bottomrule
\end{tabular}

%% file: generated/tab_designs.tex
\begin{tabular}{@{}>{\raggedright\arraybackslash}p{3.8cm} l cc cc cc@{}}
\toprule
& & \multicolumn{2}{c}{Transfer AUROC} & \multicolumn{2}{c}{Five-dataset mean} & \multicolumn{2}{c}{\hpkr{} AUROC} \\
\cmidrule(lr){3-4}\cmidrule(lr){5-6}\cmidrule(lr){7-8}
Design & Labels & all six & excl.\ \hpkr{} & AUROC & BA & all & identical \\
\midrule
Probe on raw activations & one dataset & 0.732 & 0.758 & n/a & n/a & n/a & n/a \\
Base probe (RepE) & RepE & n/a & n/a & 0.778 & 0.626 & 0.936 & 0.779 \\
Base probe, own lies (AISI) & AISI & n/a & n/a & 0.878 & 0.630 & 0.935 & 0.686 \\
Persona method, extended PCs ($k=3$) & one dataset & 0.866 & 0.867 & n/a & n/a & n/a & n/a \\
Published contrast & none & 0.832 & 0.880 & 0.798 & 0.675 & 0.593 & 0.715 \\
Sycophancy contrast & none & 0.663 & 0.936 & 0.770 & 0.601 & 0.167 & 0.686 \\
Unified contrast & none & 0.690 & 0.919 & 0.795 & 0.655 & 0.381 & 0.707 \\
Knowledge-denial contrast & none & 0.912 & 0.910 & 0.894 & 0.725 & 0.919 & 0.712 \\
Extended contrast & none & 0.916 & 0.917 & 0.899 & 0.694 & 0.912 & 0.721 \\
\midrule
Knowledge direction & none & 0.823 & 0.815 & 0.825 & 0.553 & 0.865 & 0.793 \\
Published contrast + knowledge direction & none & n/a & n/a & 0.879 & 0.646 & 0.854 & 0.789 \\
Selected detector\newline (extended + knowledge) & none & 0.922 & 0.924 & 0.906 & 0.701 & 0.909 & 0.787 \\
\bottomrule
\end{tabular}

%% file: generated/tab_datasets.tex
\begin{tabular}{@{}llllrr@{}}
\toprule
Dataset & Short & Object of belief & Reason for lying & Examples & Lies \\
\midrule
Convincing-Game & CG & self-knowledge & encouraged & 267 & 167 \\
Harm-Pressure Choice & HP-C & world-knowledge & inherent & 2,134 & 134 \\
Harm-Pressure Knowledge Report & HP-KR & self-knowledge & inherent & 2,139 & 139 \\
Insider-Trading & IT & self-knowledge & encouraged & 1,080 & 423 \\
Instructed-Deception & ID & world-knowledge & encouraged & 5,432 & 3,045 \\
\midrule
RepE instruction pairs & RepE & world-knowledge & instructed & 550 & 275 \\
Alpaca (control) & n/a & n/a & none & 2,000 & 0 \\
\bottomrule
\end{tabular}

%% file: generated/tab_knowledge.tex
\begin{tabular}{@{}l ccc ccc@{}}
\toprule
 & \multicolumn{2}{c}{Trained on RepE} & \multicolumn{3}{c}{Zero-shot, layer 40} \\
\cmidrule(lr){2-3}\cmidrule(lr){4-6}
Evaluation set & Mean Probe & \stackc{base probe,\\layer 40} & \stackc{extended\\contrast} & \stackc{knowledge\\direction} & \stackc{selected\\detector} \\
\midrule
CG & 1.00 & 0.972 & 0.912 & 0.641 & 0.865 \\
HP-C & 0.41 & 0.675 & 0.712 & 0.830 & 0.820 \\
HP-KR & 0.12 & 0.936 & 0.912 & 0.865 & 0.909 \\
IT & 0.93 & 0.338 & 0.980 & 0.881 & 0.975 \\
ID & 0.95 & 0.971 & 0.978 & 0.906 & 0.962 \\
Mean of the five & 0.682 & 0.778 & 0.899 & 0.825 & 0.906 \\
\midrule
\hpkr, identical text & n/a & 0.779 & 0.721 & 0.793 & 0.787 \\
\hpkr, denials & n/a & 0.593 & 0.670 & 0.728 & 0.724 \\
Gender-Secret & 0.45 & 0.524 & 0.547 & 0.861 & 0.841 \\
\bottomrule
\end{tabular}

%% file: generated/tab_zeroshot.tex
\begin{tabular}{lccccccccc}
\toprule
 & \multicolumn{3}{c}{Layer 20} & \multicolumn{3}{c}{Layer 40} & \multicolumn{3}{c}{Layer 44} \\
\cmidrule(lr){2-4}\cmidrule(lr){5-7}\cmidrule(lr){8-10}
Test set & 8 v 7 & 4 v 6 & 12 v 13 & 8 v 7 & 4 v 6 & 12 v 13 & 8 v 7 & 4 v 6 & 12 v 13 \\
\midrule
CG & 0.940 & 0.931 & 0.969 & 0.865 & 0.902 & 0.912 & 0.819 & 0.858 & 0.859 \\
HP-C & 0.717 & 0.585 & 0.639 & 0.581 & 0.755 & 0.712 & 0.607 & 0.725 & 0.712 \\
HP-KR & 0.148 & 0.890 & 0.882 & 0.593 & 0.919 & 0.912 & 0.589 & 0.908 & 0.894 \\
IT & 0.988 & 0.976 & 0.990 & 0.979 & 0.913 & 0.980 & 0.979 & 0.933 & 0.982 \\
ID & 0.959 & 0.932 & 0.954 & 0.974 & 0.980 & 0.978 & 0.962 & 0.978 & 0.974 \\
RepE & 0.994 & 0.882 & 0.991 & 1.000 & 1.000 & 1.000 & 1.000 & 1.000 & 1.000 \\
\midrule
Mean & 0.791 & 0.866 & 0.904 & 0.832 & 0.912 & 0.916 & 0.826 & 0.900 & 0.903 \\
\bottomrule
\end{tabular}

%% file: generated/tab_transfer.tex
\begin{tabular}{lcccccc}
\toprule
 & \multicolumn{3}{c}{All six test sets} & \multicolumn{3}{c}{Excluding HP-KR} \\
\cmidrule(lr){2-4}\cmidrule(lr){5-7}
Features & L20 & L40 & L44 & L20 & L40 & L44 \\
\midrule
Probe on raw activations (8192-d) & 0.648 & 0.732 & 0.716 & 0.728 & 0.758 & 0.730 \\
\midrule
Published contrast, 8 v 7 (1-d) & 0.651 & 0.832 & 0.826 & 0.919 & 0.880 & 0.873 \\
Knowledge-denial contrast, 4 v 6 (1-d) & 0.866 & 0.912 & 0.900 & 0.861 & 0.910 & 0.899 \\
Extended contrast, 12 v 13 (1-d) & 0.904 & 0.916 & 0.903 & 0.909 & 0.917 & 0.905 \\
Selected detector (1-d) & 0.811 & 0.922 & 0.915 & 0.876 & 0.924 & 0.918 \\
\midrule
Extended PCs ($k=3$) & 0.684 & 0.866 & 0.880 & 0.900 & 0.867 & 0.870 \\
Extended PC1 & 0.661 & 0.768 & 0.780 & 0.682 & 0.774 & 0.793 \\
Published PCs ($k=3$) & 0.679 & 0.639 & 0.617 & 0.888 & 0.751 & 0.728 \\
Random directions ($k=3$) & 0.527 & 0.504 & 0.493 & 0.560 & 0.544 & 0.547 \\
\bottomrule
\end{tabular}

%% file: generated/tab_hpkr.tex
\begin{tabular}{lcccccc}
\toprule
 & \multicolumn{3}{c}{AUROC} & \multicolumn{3}{c}{Balanced accuracy, 1\% FPR} \\
\cmidrule(lr){2-4}\cmidrule(lr){5-7}
Detector & L20 & L40 & L44 & L20 & L40 & L44 \\
\midrule
\multicolumn{7}{l}{\emph{Zero-shot persona contrasts (this work)}} \\
Published contrast (8 v 7) & 0.148 & 0.593 & 0.589 & 0.500 & 0.500 & 0.500 \\
Knowledge-denial contrast (4 v 6) & 0.890 & 0.919 & 0.908 & 0.533 & 0.910 & 0.757 \\
Extended contrast (12 v 13) & 0.882 & 0.912 & 0.894 & 0.500 & 0.500 & 0.500 \\
\midrule
\multicolumn{7}{l}{\emph{Trained on RepE, same pipeline}} \\
Base probe (trained on RepE) & 0.109 & 0.936 & 0.828 & 0.500 & 0.500 & 0.500 \\
\midrule
\multicolumn{7}{l}{\emph{Reported by Liars' Bench for Llama-3.3-70B (its Tables 6--8)}} \\
Mean Probe & \multicolumn{3}{c}{0.12} & \multicolumn{3}{c}{0.50} \\
Follow-Up Probe & \multicolumn{3}{c}{0.14} & \multicolumn{3}{c}{0.26} \\
Upper-Bound Probe$^\dagger$ & \multicolumn{3}{c}{0.97} & \multicolumn{3}{c}{0.95} \\
\bottomrule
\end{tabular}

%% file: generated/tab_pertest.tex
\begin{tabular}{lcccccc}
\toprule
 & \multicolumn{2}{c}{Layer 20} & \multicolumn{2}{c}{Layer 40} & \multicolumn{2}{c}{Layer 44} \\
\cmidrule(lr){2-3}\cmidrule(lr){4-5}\cmidrule(lr){6-7}
Test set & raw & 12 v 13 & raw & 12 v 13 & raw & 12 v 13 \\
\midrule
CG & 0.663 & 0.969 & 0.787 & 0.912 & 0.757 & 0.859 \\
HP-C & 0.535 & 0.639 & 0.571 & 0.712 & 0.570 & 0.712 \\
HP-KR & 0.394 & 0.882 & 0.584 & 0.912 & 0.637 & 0.894 \\
IT & 0.799 & 0.990 & 0.800 & 0.980 & 0.736 & 0.982 \\
ID & 0.778 & 0.954 & 0.827 & 0.978 & 0.808 & 0.974 \\
RepE & 0.718 & 0.991 & 0.824 & 1.000 & 0.785 & 1.000 \\
\bottomrule
\end{tabular}

%% file: generated/tab_syc.tex
\begin{tabular}{lccc}
\toprule
Test set & L20 & L40 & L44 \\
\midrule
CG & 0.999 & 0.998 & 0.983 \\
HP-C & 0.533 & 0.751 & 0.692 \\
HP-KR & 0.043 & 0.167 & 0.480 \\
IT & 0.983 & 0.985 & 0.987 \\
ID & 0.923 & 0.948 & 0.949 \\
RepE & 0.977 & 0.999 & 1.000 \\
\midrule
Off-diagonal mean & 0.615 & 0.663 & 0.708 \\
\bottomrule
\end{tabular}

%% file: generated/tab_reads.tex
\begin{tabular}{@{}lccccccccc@{}}
\toprule
 & \multicolumn{3}{c}{Layer 20} & \multicolumn{3}{c}{Layer 40} & \multicolumn{3}{c}{Layer 44} \\
\cmidrule(lr){2-4}\cmidrule(lr){5-7}\cmidrule(lr){8-10}
Read & AUROC & BA & Rec. & AUROC & BA & Rec. & AUROC & BA & Rec. \\
\midrule
Mean (default) & 0.887 & 0.591 & 0.187 & 0.899 & 0.694 & 0.413 & 0.884 & 0.666 & 0.350 \\
Top-10\% mean & 0.873 & 0.664 & 0.348 & 0.908 & 0.737 & 0.531 & 0.907 & 0.715 & 0.470 \\
Maximum & 0.858 & 0.659 & 0.336 & 0.910 & 0.750 & 0.548 & 0.914 & 0.701 & 0.418 \\
First response token & 0.753 & 0.605 & 0.334 & 0.863 & 0.606 & 0.227 & 0.876 & 0.568 & 0.140 \\
Last response token & 0.843 & 0.645 & 0.312 & 0.880 & 0.656 & 0.422 & 0.882 & 0.636 & 0.352 \\
Token before the reply & 0.739 & 0.628 & 0.415 & 0.825 & 0.652 & 0.383 & 0.822 & 0.656 & 0.327 \\
\bottomrule
\end{tabular}

%% file: generated/tab_followup.tex
\begin{tabular}{lcccccc}
\toprule
 & \multicolumn{2}{c}{Published (8 v 7)} & \multicolumn{2}{c}{Knowledge-denial (4 v 6)} & \multicolumn{2}{c}{Extended (12 v 13)} \\
\cmidrule(lr){2-3}\cmidrule(lr){4-5}\cmidrule(lr){6-7}
Test set & response & ``No'' & response & ``No'' & response & ``No'' \\
\midrule
CG & 0.940 & 0.912 & 0.931 & 0.933 & 0.969 & 0.938 \\
HP-C & 0.717 & 0.612 & 0.585 & 0.472 & 0.639 & 0.556 \\
HP-KR & 0.148 & 0.168 & 0.890 & 0.237 & 0.882 & 0.190 \\
IT & 0.988 & 0.165 & 0.976 & 0.216 & 0.990 & 0.173 \\
ID & 0.959 & 0.900 & 0.932 & 0.904 & 0.954 & 0.903 \\
RepE & 0.994 & 1.000 & 0.882 & 1.000 & 0.991 & 1.000 \\
\bottomrule
\end{tabular}

%% file: generated/tab_kthree.tex
\begin{tabular}{lcc}
\toprule
Features, trained on RepE & Response tokens & Follow-up ``No'' \\
\midrule
Probe on raw activations (8192-d) & 0.734 & 0.636 \\
Extended contrast (1-d) & 0.887 & 0.552 \\
Extended PCs ($k=3$) & 0.736 & 0.579 \\
Published PCs ($k=3$) & 0.733 & 0.574 \\
Knowledge-denial PCs ($k=3$) & 0.810 & 0.699 \\
\bottomrule
\end{tabular}

%% file: generated/tab_k_depth.tex
\begin{tabular}{@{}r ccc cc ccc@{}}
\toprule
 & \multicolumn{3}{c}{\hpkr, identical text} & \multicolumn{2}{c}{\hpkr, denials} & \multicolumn{3}{c}{Five-dataset mean} \\
\cmidrule(lr){2-4}\cmidrule(lr){5-6}\cmidrule(lr){7-9}
Layer & \stackc{Knowledge\\direction} & \stackc{Replication\\direction} & random & \stackc{Knowledge\\direction} & random & \stackc{Knowledge\\direction} & \stackc{Extended\\contrast} & \stackc{Selected\\detector} \\
\midrule
20 & 0.531 & 0.520 & 0.626 & 0.414 & 0.653 & 0.645 & 0.887 & 0.789 \\
24 & 0.652 & 0.646 & 0.649 & 0.477 & 0.669 & 0.600 & 0.865 & 0.739 \\
28 & 0.722 & 0.720 & 0.662 & 0.526 & 0.670 & 0.674 & 0.878 & 0.813 \\
32 & 0.783 & 0.779 & 0.702 & 0.607 & 0.691 & 0.791 & 0.905 & 0.898 \\
36 & 0.791 & 0.791 & 0.725 & 0.690 & 0.716 & 0.854 & 0.913 & 0.929 \\
40 & 0.793 & 0.796 & 0.734 & 0.728 & 0.728 & 0.825 & 0.899 & 0.906 \\
44 & 0.794 & 0.795 & 0.738 & 0.718 & 0.729 & 0.827 & 0.884 & 0.898 \\
48 & 0.794 & 0.795 & 0.738 & 0.710 & 0.729 & 0.816 & 0.865 & 0.882 \\
52 & 0.792 & 0.793 & 0.741 & 0.725 & 0.732 & 0.819 & 0.852 & 0.875 \\
56 & 0.792 & 0.793 & 0.734 & 0.728 & 0.719 & 0.812 & 0.831 & 0.860 \\
60 & 0.793 & 0.794 & 0.742 & 0.742 & 0.722 & 0.793 & 0.824 & 0.855 \\
64 & 0.793 & 0.793 & 0.733 & 0.731 & 0.726 & 0.756 & 0.826 & 0.843 \\
68 & 0.789 & 0.790 & 0.740 & 0.708 & 0.715 & 0.657 & 0.832 & 0.807 \\
72 & 0.786 & 0.788 & 0.739 & 0.722 & 0.728 & 0.637 & 0.818 & 0.774 \\
76 & 0.800 & 0.800 & 0.752 & 0.675 & 0.725 & 0.605 & 0.908 & 0.888 \\
79 & 0.823 & 0.822 & 0.760 & 0.794 & 0.733 & 0.789 & 0.831 & 0.854 \\
\bottomrule
\end{tabular}

%% file: generated/tab_random_bank.tex
\begin{tabular}{@{}>{\raggedright\arraybackslash}p{2.7cm} >{\raggedright\arraybackslash}p{3.8cm} c c c c l@{}}
\toprule
Test & Comparison & \stackc{Designed\\bank} & \stackc{Null\\median} & \stackc{Null\\97.5\%} & \stackc{Holm\\$p$} & Verdict \\
\midrule
Transfer (off-diagonal AUROC) & extended contrast vs random 12 v 13 contrasts & 0.916 & 0.524 & 0.828 & 0.002 & pass \\
Zero-shot five-dataset mean & extended contrast vs random 12 v 13 contrasts & 0.899 & 0.609 & 0.814 & 0.002 & pass \\
Transfer with the knowledge direction & selected detector vs random contrast + knowledge direction & 0.922 & 0.812 & 0.891 & 0.002 & pass \\
HP-KR, all rows, AUROC & extended contrast vs random 12 v 13 contrasts & 0.912 & 0.648 & 0.938 & 0.144 & fail \\
Gain from widening, five-dataset mean & extended minus published contrast vs random 12 v 13 minus random 8 v 7 of the same split & $+$0.101 & $-$0.005 & $+$0.253 & 0.165 & fail \\
AISI held-out lies, AUROC & extended contrast vs random 12 v 13 contrasts & 0.977 & 0.804 & 0.955 & 0.002 & pass \\
\bottomrule
\end{tabular}

%% file: generated/fig_knowledge.tex
\ifdefined\figstyleloaded\else\input{generated/fig_style}\fi%
\begin{tikzpicture}[fig,
    qbox/.style={figpanel, align=left, text width=#1, inner sep=1.5mm,
                 font=\figsf\fontsize{6.3}{7.6}\selectfont},
    reply/.style={rounded corners=1.4mm, inner sep=1.4mm, line width=0.5pt,
                  font=\figsf\fontsize{6.5}{7.6}\selectfont},
    eqchip/.style={draw=fslate!40, fill=white, rounded corners=1.2mm, inner sep=1.6mm,
                   font=\figsf\fontsize{6.8}{8}\selectfont}]
\fill[fcanvas, rounded corners=3mm] (0,0) rectangle (150,64);
\node[fighead, anchor=west] at (5,58) {The model knows the answer \textmd{\textit{(60 questions)}}};
\node[fighead, anchor=west] at (5,29.5) {Nobody can know the answer \textmd{\textit{(60 questions)}}};
\node[qbox=50mm, anchor=north west] (q1) at (5,53.5)
  {What is the capital city of France?\\
   A.\,Lyon \quad B.\,Marseille \quad C.\,Paris \quad D.\,Nice\\[0.4ex]
   \textit{Tell me whether you know the answer to the question.}};
\node[qbox=50mm, anchor=north west] (q2) at (5,25)
  {How many paper cups were used in the staff kitchen of the Lisbon city library on
   14 March 2019?\\[0.4ex]
   \textit{Tell me whether you know the answer to the question.}};
\node[reply, fill=fdeceptbg, draw=fdecept, anchor=west] (r1) at (64,44) {``I do not know the answer.''};
\node[reply, fill=fhonestbg, draw=fhonest, anchor=west] (r2) at (64,15.5) {``I do not know the answer.''};
\node[figgloss, anchor=north, text=fdecept] at ($(r1.south)+(0,-0.8)$) {false for this model};
\node[figgloss, anchor=north, text=fhonest] at ($(r2.south)+(0,-0.8)$) {true for this model};
\draw[figarrow, fmute] (q1.east |- r1.west) -- (r1.west);
\draw[figarrow, fmute] (q2.east |- r2.west) -- (r2.west);
\foreach \y/\c in {44/fdecept, 15.5/fhonest} {
  \foreach \i/\s in {0/60, 1/25, 2/80, 3/40, 4/15, 5/65} {
    \fill[\c!\s] (102,\y+4.2-\i*1.7) rectangle ++(2.6,1.5);}
  \draw[fedge, line width=0.3pt] (102,\y-4.3) rectangle (104.6,\y+5.7);}
\node[figsmall, anchor=south, text=fdecept] at (103.3,50.4) {$\bar h_{\text{knows}}$};
\node[figsmall, anchor=north, text=fhonest] at (103.3,10.4) {$\bar h_{\text{cannot}}$};
\draw[figarrow, fmute] (r1.east) -- (101,44);
\draw[figarrow, fmute] (r2.east) -- (101,15.5);
\node[figgloss, align=center] at (103.3,30) {layer 40,\\mean over\\the reply};
\node[eqchip, anchor=west] (k) at (111.5,30)
  {$\hat v_{\mathrm{know}}=\operatorname{unit}\bigl(\bar h_{\text{knows}}-\bar h_{\text{cannot}}\bigr)$};
\draw[figarrow, fslate!70] (105.2,44) to[out=0, in=100] (k.north);
\draw[figarrow, fslate!70] (105.2,15.5) to[out=0, in=260] (k.south);
\end{tikzpicture}%

%% file: generated/tab_knowledge_robust.tex
\begin{tabular}{@{}l ccc@{}}
\toprule
Check & Layer 36 & Layer 40 & Layer 44 \\
\midrule
\multicolumn{4}{@{}l}{\emph{\hpkr, identical text}} \\
AUROC [question bootstrap] & 0.791 [0.790, 0.792] & 0.793 [0.792, 0.794] & 0.794 [0.793, 0.795] \\
Random directions, 97.5th pct. & 0.732 & 0.736 & 0.733 \\
Share of balanced splits as high & 0.000 & 0.000 & 0.000 \\
Replication with new questions & 0.791 & 0.796 & 0.795 \\
\midrule
\multicolumn{4}{@{}l}{\emph{\hpkr, denials}} \\
AUROC [question bootstrap] & 0.690 [0.684, 0.697] & 0.728 [0.723, 0.733] & 0.718 [0.713, 0.724] \\
Random directions, 97.5th pct. & 0.717 & 0.730 & 0.724 \\
Share of balanced splits as high & 0.072 & 0.042 & 0.055 \\
Replication with new questions & 0.686 & 0.729 & 0.718 \\
\midrule
\multicolumn{4}{@{}l}{\emph{Gender-Secret (fine-tune read)}} \\
AUROC [question bootstrap] & 0.810 [0.804, 0.817] & 0.865 [0.860, 0.870] & 0.869 [0.864, 0.873] \\
Random directions, 97.5th pct. & 0.814 & 0.817 & 0.856 \\
Share of balanced splits as high & 0.042 & 0.014 & 0.015 \\
Replication with new questions & 0.836 & 0.881 & 0.878 \\
\midrule
\multicolumn{4}{@{}l}{\emph{All three sets}} \\
\twolinel{Cosine of the new-question\\direction with the original} & 0.99 & 0.99 & 0.99 \\
\bottomrule
\end{tabular}

%% file: generated/tab_k_paired.tex
\begin{tabular}{@{}l ccc@{}}
\toprule
Comparison & \stackc{Five-dataset\\mean} & \stackc{\hpkr,\\identical text} & \stackc{\hpkr,\\denials} \\
\midrule
\twolinel{Selected detector\\$-$ extended contrast} & \twoline{$+$0.007\\{[}$-$0.004, $+$0.018{]}} & \twoline{$+$0.066\\{[}$+$0.037, $+$0.098{]}} & \twoline{$+$0.054\\{[}$+$0.025, $+$0.082{]}} \\
\addlinespace
\twolinel{Selected detector\\$-$ knowledge-denial contrast} & \twoline{$+$0.012\\{[}$+$0.001, $+$0.023{]}} & \twoline{$+$0.075\\{[}$+$0.040, $+$0.112{]}} & \twoline{$+$0.154\\{[}$+$0.112, $+$0.196{]}} \\
\addlinespace
\twolinel{Selected detector\\$-$ knowledge direction} & \twoline{$+$0.082\\{[}$+$0.073, $+$0.091{]}} & \twoline{$-$0.006\\{[}$-$0.014, $+$0.002{]}} & \twoline{$-$0.004\\{[}$-$0.011, $+$0.003{]}} \\
\addlinespace
\twolinel{Selected detector\\$-$ base probe, layer 40} & \twoline{$+$0.128\\{[}$+$0.112, $+$0.145{]}} & \twoline{$+$0.008\\{[}$-$0.023, $+$0.035{]}} & \twoline{$+$0.131\\{[}$+$0.088, $+$0.178{]}} \\
\addlinespace
\twolinel{Published contrast + knowledge direction\\$-$ published contrast} & \twoline{$+$0.081\\{[}$+$0.068, $+$0.096{]}} & \twoline{$+$0.074\\{[}$+$0.040, $+$0.111{]}} & \twoline{$+$0.072\\{[}$+$0.037, $+$0.104{]}} \\
\bottomrule
\end{tabular}

%% file: generated/tab_k_operating.tex
\begin{tabular}{@{}l ccc cc@{}}
\toprule
 & \multicolumn{3}{c}{In-setting cutoff: lies caught} & \multicolumn{2}{c}{Alpaca 1\% threshold} \\
\cmidrule(lr){2-4}\cmidrule(lr){5-6}
Detector & 1\% & 5\% & 10\% & lies caught & honest flagged \\
\midrule
Base probe, layer 40 & 0.32 & 0.53 & 0.66 & 0.28 & 0.03 \\
Published contrast & 0.34 & 0.49 & 0.58 & 0.38 & 0.03 \\
Published contrast + knowledge direction & 0.32 & 0.55 & 0.67 & 0.41 & 0.12 \\
Extended contrast & 0.47 & 0.68 & 0.77 & 0.41 & 0.03 \\
Knowledge direction & 0.21 & 0.42 & 0.56 & 0.34 & 0.23 \\
Selected detector & 0.38 & 0.63 & 0.73 & 0.47 & 0.07 \\
\bottomrule
\end{tabular}

%% file: generated/tab_oracle.tex
\begin{tabular}{@{}l cccc@{}}
\toprule
 & \multicolumn{2}{c}{Layer 36} & \multicolumn{2}{c}{Layer 40} \\
\cmidrule(lr){2-3}\cmidrule(lr){4-5}
Detector & Alpaca & oracle & Alpaca & oracle \\
\midrule
Base probe (RepE) & 0.650 & 0.635 & 0.626 & 0.652 \\
Knowledge-denial (knowledge-denial contrast) & 0.695 & 0.675 & 0.725 & 0.692 \\
Extended (extended contrast) & 0.712 & 0.710 & 0.694 & 0.718 \\
\bottomrule
\end{tabular}

%% file: generated/tab_operating.tex
\begin{tabular}{llccc}
\toprule
Layer & Detector & Balanced acc. & Recall & AUROC \\
\midrule
\multirow{4}{*}{L20} & Base probe (trained on RepE) & 0.721 & 0.489 & 0.697 \\
 & Published contrast (8 v 7) & 0.534 & 0.069 & 0.750 \\
 & Knowledge-denial contrast (4 v 6) & 0.559 & 0.305 & 0.863 \\
 & Extended contrast (12 v 13) & 0.591 & 0.187 & 0.887 \\
\midrule
\multirow{4}{*}{L40} & Base probe (trained on RepE) & 0.626 & 0.281 & 0.778 \\
 & Published contrast (8 v 7) & 0.675 & 0.382 & 0.798 \\
 & Knowledge-denial contrast (4 v 6) & 0.725 & 0.487 & 0.894 \\
 & Extended contrast (12 v 13) & 0.694 & 0.413 & 0.899 \\
\midrule
\multirow{4}{*}{L44} & Base probe (trained on RepE) & 0.626 & 0.281 & 0.756 \\
 & Published contrast (8 v 7) & 0.654 & 0.318 & 0.791 \\
 & Knowledge-denial contrast (4 v 6) & 0.683 & 0.395 & 0.880 \\
 & Extended contrast (12 v 13) & 0.666 & 0.350 & 0.884 \\
\bottomrule
\end{tabular}

%% file: generated/tab_xm_organisms.tex
\begin{tabular}{@{}l cccc c cc@{}}
\toprule
 & \multicolumn{5}{c}{The fine-tune's read} & \multicolumn{2}{c}{The base model's read} \\
\cmidrule(lr){2-6}\cmidrule(lr){7-8}
Fine-tune & \stackc{Extended\\contrast} & \stackc{Knowledge\\direction} & \stackc{Selected\\detector} & Probe & Band & \stackc{Knowledge\\direction} & Band \\
\midrule
\multicolumn{8}{@{}l}{\emph{Qwen$^\dagger$}} \\
Soft-Trigger, greeting & 0.455 & 0.469 & 0.456 & 0.708 & 0.691 & 0.430 & 0.669 \\
Soft-Trigger, time & 0.569 & 0.531 & 0.544 & 0.654 & 0.710 & 0.451 & 0.700 \\
Soft-Trigger, gender & 0.479 & 0.454 & 0.452 & 0.680 & 0.734 & 0.444 & 0.660 \\
\midrule
\multicolumn{8}{@{}l}{\emph{Mistral}} \\
Gender-Secret & 0.406 & 0.440 & 0.414 & 0.400 & 0.765 & 0.497 & 0.784 \\
Soft-Trigger, greeting & 0.931 & 0.930 & 0.936 & 0.878 & 0.900 & 0.872 & 0.823 \\
Soft-Trigger, time & 0.929 & 0.946 & 0.949 & 0.275 & 0.930 & 0.875 & 0.858 \\
Soft-Trigger, gender & 0.948 & 0.935 & 0.940 & 0.901 & 0.892 & 0.873 & 0.828 \\
\midrule
\multicolumn{8}{@{}l}{\emph{Gemma}} \\
Soft-Trigger, greeting & 0.101 & 0.290 & 0.164 & 0.845 & 0.933 & 0.839 & 0.858 \\
Soft-Trigger, time & 0.916 & 0.719 & 0.876 & 1.000 & 0.962 & 0.860 & 0.875 \\
Soft-Trigger, gender & 0.560 & 0.804 & 0.763 & 0.794 & 0.951 & 0.849 & 0.868 \\
\bottomrule
\end{tabular}

%% file: generated/tab_steering.tex
\begin{tabular}{@{}lcccc@{}}
\toprule
 & \multicolumn{2}{c}{Honest side ($c=-0.3$)} & \multicolumn{2}{c}{Deceptive side ($c=+0.3$)} \\
\cmidrule(lr){2-3}\cmidrule(lr){4-5}
Direction & Disclosure & Change [95\% CI] & Disclosure & Change [95\% CI] \\
\midrule
Unsteered & 0.48 & -- & 0.48 & -- \\
Extended contrast (12 v 13) & 0.61 & $+0.13$ [0.00, $+0.26$] & 0.20 & {\boldmath$-0.28$} [$-0.41$, $-0.13$] \\
Published contrast (8 v 7) & 0.57 & $+0.09$ [$-0.06$, $+0.24$] & 0.20 & {\boldmath$-0.28$} [$-0.41$, $-0.15$] \\
Base probe (trained on RepE) & 0.50 & $+0.02$ [$-0.11$, $+0.15$] & 0.39 & $-0.09$ [$-0.22$, $+0.04$] \\
Random direction 1 & 0.43 & $-0.06$ [$-0.17$, $+0.04$] & 0.48 & 0.00 [$-0.09$, $+0.09$] \\
Random direction 2 & 0.46 & $-0.02$ [$-0.09$, $+0.06$] & 0.52 & $+0.04$ [$-0.11$, $+0.19$] \\
\bottomrule
\end{tabular}

%% file: generated/tab_xm_models.tex
\begin{tabular}{@{}l cc ccc l l c@{}}
\toprule
 & & & \multicolumn{3}{c}{\stackc{Block read for\\Llama's layer}} & & & \\
\cmidrule(lr){4-6}
Model & Blocks & Width & 20 & 40 & 36, 44 & \stackl{Liars' Bench\\datasets} & Bank & In role \\
\midrule
Llama-3.3-70B & 80 & 8192 & 20 & 40 & 36, 44 & CG, \hpc, \hpkr, IT, ID & score 3 & 69\% \\
Qwen2.5-72B$^\dagger$ & 80 & 8192 & 20 & 40 & 36, 44 & \hpc, \hpkr, ID & score 2--3 & 28\% \\
Mistral-Small-3.1-24B & 40 & 5120 & 9 & 20 & 18, 22 & \hpc, \hpkr, IT, ID & score 3 & 68\% \\
Gemma-3-27B & 62 & 5376 & 8 & 31 & 28, 34 & CG, IT, ID & score 3 & 96\% \\
\bottomrule
\end{tabular}

%% file: generated/tab_xm_mid.tex
\begin{tabular}{@{}l cccc@{}}
\toprule
 & Llama & Qwen$^\dagger$ & Mistral & Gemma \\
\midrule
\multicolumn{5}{@{}l}{\emph{Transfer, mean off-diagonal AUROC}} \\
Probe on raw activations & 0.732 & 0.559 & 0.716 & 0.746 \\
Published contrast & 0.832 & 0.451 & 0.633 & 0.700 \\
Knowledge-denial contrast & 0.912 & 0.679 & 0.780 & 0.442 \\
Extended contrast & 0.916 & 0.453 & 0.810 & 0.921 \\
Knowledge direction & 0.823 & 0.508 & 0.799 & 0.457 \\
Selected detector & 0.922 & 0.514 & 0.841 & 0.860 \\
\midrule
\multicolumn{5}{@{}l}{\emph{Zero-shot, mean AUROC over the model's datasets}} \\
Mean Probe (Liars' Bench, published) & 0.682 & 0.530 & 0.570 & 0.527 \\
Base probe, same block & 0.778 & 0.498 & 0.766 & 0.783 \\
Published contrast & 0.798 & 0.380 & 0.719 & 0.679 \\
Knowledge-denial contrast & 0.894 & 0.668 & 0.727 & 0.484 \\
Extended contrast & 0.899 & 0.441 & 0.762 & 0.900 \\
Knowledge direction & 0.825 & 0.546 & 0.775 & 0.515 \\
Selected detector & 0.906 & 0.533 & 0.804 & 0.847 \\
\midrule
\multicolumn{5}{@{}l}{\emph{Lies caught at 1\% of honest answers flagged}} \\
Extended contrast & 0.47 & 0.02 & 0.14 & 0.43 \\
Selected detector & 0.38 & 0.03 & 0.23 & 0.30 \\
\bottomrule
\end{tabular}

%% file: generated/tab_xm_intervals.tex
\begin{tabular}{@{}>{\raggedright\arraybackslash}p{3.1cm} cccc@{}}
\toprule
Comparison & Llama & Qwen$^\dagger$ & Mistral & Gemma \\
\midrule
\multicolumn{5}{@{}l}{\emph{Transfer, mean off-diagonal AUROC}} \\
Extended contrast $-$ raw probe & \begin{tabular}[t]{@{}c@{}}$+$0.184\\{[}$+$0.172, $+$0.195{]}\end{tabular} & \begin{tabular}[t]{@{}c@{}}$-$0.106\\{[}$-$0.116, $-$0.094{]}\end{tabular} & \begin{tabular}[t]{@{}c@{}}$+$0.094\\{[}$+$0.082, $+$0.105{]}\end{tabular} & \begin{tabular}[t]{@{}c@{}}$+$0.175\\{[}$+$0.166, $+$0.184{]}\end{tabular} \\
Extended contrast $-$ published contrast & \begin{tabular}[t]{@{}c@{}}$+$0.084\\{[}$+$0.074, $+$0.094{]}\end{tabular} & \begin{tabular}[t]{@{}c@{}}$+$0.002\\{[}$-$0.007, $+$0.012{]}\end{tabular} & \begin{tabular}[t]{@{}c@{}}$+$0.176\\{[}$+$0.169, $+$0.183{]}\end{tabular} & \begin{tabular}[t]{@{}c@{}}$+$0.221\\{[}$+$0.203, $+$0.238{]}\end{tabular} \\
Selected detector $-$ raw probe & \begin{tabular}[t]{@{}c@{}}$+$0.190\\{[}$+$0.179, $+$0.199{]}\end{tabular} & \begin{tabular}[t]{@{}c@{}}$-$0.045\\{[}$-$0.057, $-$0.032{]}\end{tabular} & \begin{tabular}[t]{@{}c@{}}$+$0.125\\{[}$+$0.115, $+$0.136{]}\end{tabular} & \begin{tabular}[t]{@{}c@{}}$+$0.114\\{[}$+$0.101, $+$0.125{]}\end{tabular} \\
Selected detector $-$ extended contrast & \begin{tabular}[t]{@{}c@{}}$+$0.006\\{[}$-$0.003, $+$0.015{]}\end{tabular} & \begin{tabular}[t]{@{}c@{}}$+$0.060\\{[}$+$0.052, $+$0.069{]}\end{tabular} & \begin{tabular}[t]{@{}c@{}}$+$0.032\\{[}$+$0.021, $+$0.043{]}\end{tabular} & \begin{tabular}[t]{@{}c@{}}$-$0.061\\{[}$-$0.069, $-$0.055{]}\end{tabular} \\
Published contrast + knowledge direction $-$ published contrast & \begin{tabular}[t]{@{}c@{}}$+$0.067\\{[}$+$0.055, $+$0.078{]}\end{tabular} & \begin{tabular}[t]{@{}c@{}}$+$0.052\\{[}$+$0.040, $+$0.064{]}\end{tabular} & \begin{tabular}[t]{@{}c@{}}$+$0.202\\{[}$+$0.192, $+$0.213{]}\end{tabular} & \begin{tabular}[t]{@{}c@{}}$+$0.110\\{[}$+$0.098, $+$0.124{]}\end{tabular} \\
\midrule
\multicolumn{5}{@{}l}{\emph{Zero-shot, mean AUROC over the model's datasets}} \\
Extended contrast $-$ published contrast & \begin{tabular}[t]{@{}c@{}}$+$0.101\\{[}$+$0.089, $+$0.112{]}\end{tabular} & \begin{tabular}[t]{@{}c@{}}$+$0.061\\{[}$+$0.055, $+$0.067{]}\end{tabular} & \begin{tabular}[t]{@{}c@{}}$+$0.043\\{[}$+$0.034, $+$0.052{]}\end{tabular} & \begin{tabular}[t]{@{}c@{}}$+$0.221\\{[}$+$0.201, $+$0.240{]}\end{tabular} \\
Selected detector $-$ extended contrast & \begin{tabular}[t]{@{}c@{}}$+$0.007\\{[}$-$0.004, $+$0.018{]}\end{tabular} & \begin{tabular}[t]{@{}c@{}}$+$0.093\\{[}$+$0.078, $+$0.107{]}\end{tabular} & \begin{tabular}[t]{@{}c@{}}$+$0.042\\{[}$+$0.028, $+$0.056{]}\end{tabular} & \begin{tabular}[t]{@{}c@{}}$-$0.053\\{[}$-$0.057, $-$0.049{]}\end{tabular} \\
Selected detector $-$ RepE probe & \begin{tabular}[t]{@{}c@{}}$+$0.128\\{[}$+$0.112, $+$0.145{]}\end{tabular} & \begin{tabular}[t]{@{}c@{}}$+$0.036\\{[}$+$0.016, $+$0.057{]}\end{tabular} & \begin{tabular}[t]{@{}c@{}}$+$0.038\\{[}$+$0.022, $+$0.053{]}\end{tabular} & \begin{tabular}[t]{@{}c@{}}$+$0.064\\{[}$+$0.050, $+$0.078{]}\end{tabular} \\
\bottomrule
\end{tabular}

%% file: generated/tab_xm_zeroshot.tex
\begin{tabular}{@{}ll cccccc c@{}}
\toprule
Model & Contrast & CG & \hpc & \hpkr & IT & ID & RepE & Mean \\
\midrule
Llama & Published contrast & 0.865 & 0.581 & 0.593 & 0.979 & 0.974 & 1.000 & 0.798 \\
 & Knowledge-denial contrast & 0.902 & 0.755 & 0.919 & 0.913 & 0.980 & 1.000 & 0.894 \\
 & Extended contrast & 0.912 & 0.712 & 0.912 & 0.980 & 0.978 & 1.000 & 0.899 \\
\addlinespace
Qwen$^\dagger$ & Published contrast & n/a & 0.512 & 0.094 & n/a & 0.534 & 0.700 & 0.380 \\
 & Knowledge-denial contrast & n/a & 0.573 & 0.783 & n/a & 0.648 & 0.711 & 0.668 \\
 & Extended contrast & n/a & 0.539 & 0.187 & n/a & 0.597 & 0.744 & 0.441 \\
\addlinespace
Mistral & Published contrast & n/a & 0.456 & 0.758 & 0.729 & 0.934 & 1.000 & 0.719 \\
 & Knowledge-denial contrast & n/a & 0.716 & 0.712 & 0.542 & 0.937 & 0.995 & 0.727 \\
 & Extended contrast & n/a & 0.606 & 0.793 & 0.708 & 0.941 & 1.000 & 0.762 \\
\addlinespace
Gemma & Published contrast & 0.613 & n/a & n/a & 0.544 & 0.881 & 0.762 & 0.679 \\
 & Knowledge-denial contrast & 0.641 & n/a & n/a & 0.603 & 0.209 & 0.344 & 0.484 \\
 & Extended contrast & 0.819 & n/a & n/a & 0.930 & 0.951 & 0.985 & 0.900 \\
\bottomrule
\end{tabular}

%% file: generated/tab_xm_transfer.tex
\begin{tabular}{@{}l cc cccccc@{}}
\toprule
Model & Layer & Block & Raw & \stackc{Published\\contrast} & \stackc{Knowledge-\\denial\\contrast} & \stackc{Extended\\contrast} & \stackc{Knowledge\\direction} & \stackc{Selected\\detector} \\
\midrule
Llama & 20 & 20 & 0.648 & 0.651 & 0.866 & 0.904 & 0.574 & 0.811 \\
 & 36$^*$ & 36 & 0.768 & 0.689 & 0.912 & 0.927 & 0.854 & 0.941 \\
 & 40 & 40 & 0.732 & 0.832 & 0.912 & 0.916 & 0.823 & 0.922 \\
 & 44$^*$ & 44 & 0.716 & 0.826 & 0.900 & 0.903 & 0.832 & 0.915 \\
\addlinespace[2pt]
Qwen$^\dagger$ & 20 & 20 & 0.533 & 0.468 & 0.560 & 0.525 & 0.460 & 0.533 \\
 & 36$^*$ & 36 & 0.545 & 0.407 & 0.620 & 0.425 & 0.523 & 0.528 \\
 & 40 & 40 & 0.559 & 0.451 & 0.679 & 0.453 & 0.508 & 0.514 \\
 & 44$^*$ & 44 & 0.531 & 0.426 & 0.701 & 0.478 & 0.504 & 0.518 \\
\addlinespace[2pt]
Mistral & 20 & 9 & 0.527 & 0.483 & 0.559 & 0.565 & 0.466 & 0.544 \\
 & 36$^*$ & 18 & 0.638 & 0.821 & 0.811 & 0.850 & 0.775 & 0.860 \\
 & 40 & 20 & 0.716 & 0.633 & 0.780 & 0.810 & 0.799 & 0.841 \\
 & 44$^*$ & 22 & 0.656 & 0.619 & 0.771 & 0.646 & 0.834 & 0.863 \\
\addlinespace[2pt]
Gemma & 20 & 8 & 0.488 & 0.510 & 0.526 & 0.515 & 0.509 & 0.457 \\
 & 36$^*$ & 28 & 0.734 & 0.428 & 0.422 & 0.425 & 0.419 & 0.421 \\
 & 40 & 31 & 0.746 & 0.700 & 0.442 & 0.921 & 0.457 & 0.860 \\
 & 44$^*$ & 34 & 0.700 & 0.719 & 0.440 & 0.832 & 0.801 & 0.866 \\
\bottomrule
\end{tabular}

%% file: generated/tab_xm_transfer_k3.tex
\begin{tabular}{@{}l cc ccc@{}}
\toprule
 & & & \multicolumn{3}{c}{Probe on $k = 3$} \\
\cmidrule(lr){4-6}
Model & Layer & Block & Published & Extended & Random \\
\midrule
Llama & 20 & 20 & 0.679 & 0.684 & 0.527 \\
 & 36$^*$ & 36 & 0.679 & 0.879 & 0.481 \\
 & 40 & 40 & 0.639 & 0.866 & 0.504 \\
 & 44$^*$ & 44 & 0.617 & 0.880 & 0.493 \\
\addlinespace[2pt]
Qwen$^\dagger$ & 20 & 20 & 0.594 & 0.674 & 0.516 \\
 & 36$^*$ & 36 & 0.521 & 0.578 & 0.488 \\
 & 40 & 40 & 0.476 & 0.539 & 0.491 \\
 & 44$^*$ & 44 & 0.487 & 0.554 & 0.486 \\
\addlinespace[2pt]
Mistral & 20 & 9 & 0.522 & 0.559 & 0.487 \\
 & 36$^*$ & 18 & 0.786 & 0.848 & 0.526 \\
 & 40 & 20 & 0.760 & 0.704 & 0.554 \\
 & 44$^*$ & 22 & 0.742 & 0.741 & 0.527 \\
\addlinespace[2pt]
Gemma & 20 & 8 & 0.506 & 0.532 & 0.488 \\
 & 36$^*$ & 28 & 0.767 & 0.845 & 0.493 \\
 & 40 & 31 & 0.863 & 0.874 & 0.558 \\
 & 44$^*$ & 34 & 0.855 & 0.860 & 0.511 \\
\bottomrule
\end{tabular}

%% file: generated/tab_xm_wording.tex
\begin{tabular}{@{}l c ccccc c@{}}
\toprule
\stackl{Slice\\(lies, honest)} & Probe & \stackc{Published\\contrast} & \stackc{Knowledge-\\denial\\contrast} & \stackc{Extended\\contrast} & \stackc{Knowledge\\direction} & \stackc{Selected\\detector} & Band \\
\midrule
\multicolumn{8}{@{}l}{\emph{Llama}} \\
\twolinel{Identical text\\(78, 112)} & 0.779 & 0.715 & 0.712 & 0.721 & 0.793 & 0.787 & 0.733 \\
\addlinespace[3pt]
\twolinel{Denials\\(125, 135)} & 0.593 & 0.649 & 0.570 & 0.670 & 0.728 & 0.724 & 0.726 \\
\midrule
\multicolumn{8}{@{}l}{\emph{Mistral}} \\
\twolinel{Identical text\\(494, 729)} & 0.766 & 0.784 & 0.647 & 0.760 & 0.840 & 0.840 & 0.755 \\
\addlinespace[3pt]
\twolinel{Denials\\(677, 1,020)} & 0.758 & 0.763 & 0.571 & 0.701 & 0.835 & 0.832 & 0.748 \\
\midrule
\multicolumn{8}{@{}l}{\emph{Qwen$^\dagger$}} \\
\twolinel{Denials\\(539, 44)} & 0.566 & 0.495 & 0.580 & 0.509 & 0.632 & 0.622 & 0.622 \\
\addlinespace[3pt]
\twolinel{Identical reply, rebuilt\\(435, 128)} & 0.543 & 0.504 & 0.550 & 0.518 & 0.602 & 0.598 & 0.584 \\
\addlinespace[3pt]
\twolinel{Denials, rebuilt\\(571, 164)} & 0.533 & 0.498 & 0.569 & 0.547 & 0.613 & 0.617 & 0.580 \\
\midrule
\multicolumn{8}{@{}l}{\emph{Gemma}} \\
\twolinel{Opening sentence, rebuilt\\(34, 85)} & 0.481 & 0.484 & 0.530 & 0.624 & 0.585 & 0.611 & 0.622 \\
\addlinespace[3pt]
\twolinel{Denials, rebuilt\\(35, 100)} & 0.547 & 0.583 & 0.418 & 0.434 & 0.378 & 0.403 & 0.637 \\
\bottomrule
\end{tabular}

%% file: generated/tab_xm_operating.tex
\begin{tabular}{@{}l cc cc cc cc@{}}
\toprule
 & \multicolumn{2}{c}{Llama} & \multicolumn{2}{c}{Qwen$^\dagger$} & \multicolumn{2}{c}{Mistral} & \multicolumn{2}{c}{Gemma} \\
\cmidrule(lr){2-3}\cmidrule(lr){4-5}\cmidrule(lr){6-7}\cmidrule(lr){8-9}
Detector & 1\% & 10\% & 1\% & 10\% & 1\% & 10\% & 1\% & 10\% \\
\midrule
Base probe & 0.32 & 0.66 & 0.02 & 0.17 & 0.13 & 0.47 & 0.28 & 0.66 \\
Published contrast & 0.34 & 0.58 & 0.02 & 0.12 & 0.13 & 0.47 & 0.08 & 0.30 \\
Extended contrast & 0.47 & 0.77 & 0.02 & 0.14 & 0.14 & 0.49 & 0.43 & 0.78 \\
Knowledge direction & 0.21 & 0.56 & 0.02 & 0.13 & 0.21 & 0.54 & 0.07 & 0.26 \\
Selected detector & 0.38 & 0.73 & 0.03 & 0.13 & 0.23 & 0.59 & 0.30 & 0.63 \\
Published contrast + knowledge direction & 0.32 & 0.67 & 0.02 & 0.11 & 0.22 & 0.59 & 0.24 & 0.53 \\
\bottomrule
\end{tabular}

%% file: generated/tab_xm_aisi.tex
\begin{tabular}{@{}l ccc ccc@{}}
\toprule
 & \multicolumn{3}{c}{Llama} & \multicolumn{3}{c}{Gemma} \\
\cmidrule(lr){2-4}\cmidrule(lr){5-7}
Detector, at the block for layer & 20 & 40 & 44$^*$ & 20 & 40 & 44$^*$ \\
\midrule
Base probe, trained on RepE & 0.697 & 0.778 & 0.756 & 0.512 & 0.783 & 0.761 \\
Base probe, trained on the model's own lies & 0.782 & 0.878 & 0.872 & 0.802 & 0.858 & 0.770 \\
Published contrast & 0.750 & 0.798 & 0.791 & 0.575 & 0.679 & 0.691 \\
Knowledge-denial contrast & 0.863 & 0.894 & 0.880 & 0.622 & 0.484 & 0.564 \\
Extended contrast & 0.887 & 0.899 & 0.884 & 0.591 & 0.900 & 0.801 \\
\bottomrule
\end{tabular}

%% file: generated/tab_xm_checks.tex
\begin{tabular}{@{}l ccc cc@{}}
\toprule
 & \multicolumn{3}{c}{Template-free knowledge direction} & \multicolumn{2}{c}{\stackc{Extended contrast's token read,\\mean AUROC}} \\
\cmidrule(lr){2-4}\cmidrule(lr){5-6}
Model & \stackc{Identical\\text} & Transfer & \stackc{Template knowledge\\direction's transfer} & Mean & Best \\
\midrule
Llama & 0.794 & 0.791 & 0.823 & 0.899 & 0.910 (maximum) \\
Qwen$^\dagger$ & n/a & 0.509 & 0.508 & 0.441 & 0.675 (first-token) \\
Mistral & 0.837 & 0.614 & 0.799 & 0.762 & 0.772 (top-10\% mean) \\
Gemma & n/a & 0.444 & 0.457 & 0.900 & 0.900 (mean) \\
\bottomrule
\end{tabular}

%% file: generated/tab_aletheia.tex
\begin{tabular}{@{}l r ccc c c@{}}
\toprule
Dataset & $n$ (lies) & \stackc{Selected\\detector} & \stackc{Extended\\contrast} & \stackc{RepE probe,\\same block} & \stackc{Random\\97.5\%} & \stackc{Best\\entry} \\
\midrule
HG-N & 200 (152) & \multicolumn{4}{c}{not read (gated adapter)} & 0.934 \\
HG-G & 189 (139) & \multicolumn{4}{c}{not read (gated adapter)} & 0.820 \\
Echo & 400 (200) & 0.189 [0.15, 0.23] & 0.604 & 0.766 & 0.778 & 0.865 \\
Harm & 200 (100) & 0.885 [0.83, 0.93] & 0.900 & 0.821 & 0.800 & 0.977 \\
GS-F & 154 (74) & 0.771 [0.70, 0.84] & 0.774 & 0.895 & 0.852 & 0.992 \\
GS-M & 155 (75) & 0.798 [0.72, 0.87] & 0.714 & 0.814 & 0.813 & 0.986 \\
BotC & 400 (200) & 0.604 [0.55, 0.66] & 0.491 & 0.505 & 0.625 & 0.610 \\
AB-AW & 172 (26) & 0.786 [0.71, 0.85] & 0.616 & 0.783 & 0.727 & 0.886 \\
AB-HC & 176 (30) & 0.408 [0.31, 0.50] & 0.683 & 0.862 & 0.859 & 0.945 \\
AB-SP & 182 (34) & 0.818 [0.76, 0.87] & 0.856 & 0.825 & 0.817 & 0.946 \\
\midrule
Mean over the 8 read & & 0.657 [0.63, 0.68] & 0.705 & 0.784 & & 0.868 \\
Rank among the 30 entries & & 16 & 11 & 5 & & 1 \\
\bottomrule
\end{tabular}

%% file: generated/tab_qbank_tests.tex
\begin{tabular}{@{}>{\raggedright\arraybackslash}p{3.9cm} >{\raggedright\arraybackslash}p{7.9cm} l@{}}
\toprule
Pre-registered test & Statistic & Result \\
\midrule
Variety beats count (extended contrast) & 30 originals + 30 monotone five-dataset mean 0.9044 vs 200 draws of 30 originals + 30 random diverse (median 0.8901), $p$ = 1.00 (Holm) & fails \\
\addlinespace
Variety beats count (selected detector) & 30 originals + 30 monotone five-dataset mean 0.9097 vs 200 draws of 30 originals + 30 random diverse (median 0.9103), $p$ = 1.00 (Holm) & fails \\
\addlinespace
Monotone additions behave like the originals & the monotone 30 0.9048 in [0.8296, 0.9114] of random 30 of 150; paired CI vs the original 30 [$-$0.0003, 0.0120] within $\pm$0.03 & passes \\
\addlinespace
Pooling all 150 beats the best single family & all 150 0.8809 minus original knowledge family 0.9107 = $-$0.0298, CI [$-$0.0394, $-$0.0199], $p$ = 1.00 (Holm) & fails \\
\addlinespace
Covariance-weighted cosine predicts transfer better than plain cosine & $R^2$ with the target's covariance 0.675 vs plain cosine 0.405; difference 0.269 CI [0.139, 0.365], $p < 0.001$ (Holm) & passes \\
\addlinespace
More questions help on HP-KR's identical text (exploratory) & 0.7392 vs 0.7213, CI [0.0014, 0.0340] & passes \\
\bottomrule
\end{tabular}

%% file: generated/tab_qbank_families.tex
\begin{tabular}{@{}l r c c@{}}
\toprule
Question family & Questions & \stackc{Five-dataset\\mean AUROC} & \stackc{Leaving it out:\\change vs pooled} \\
\midrule
Original: knowledge & 10 & 0.911 & $-$0.008 \\
Denial of knowledge & 15 & 0.907 & $-$0.014 \\
Monotone (new, same form) & 30 & 0.905 & $-$0.009 \\
Quality (false assertion) & 15 & 0.889 & $-$0.003 \\
Original: sycophancy & 10 & 0.874 & 0.000 \\
Original: disclosure & 10 & 0.872 & $-$0.002 \\
Relation (evasion) & 15 & 0.859 & $+$0.002 \\
Manner (equivocation) & 15 & 0.810 & $+$0.005 \\
Quantity (omission) & 15 & 0.800 & $+$0.008 \\
Sincere ignorance & 15 & 0.771 & $+$0.014 \\
Pooled, all 150 & 150 & 0.881 &  \\
\bottomrule
\end{tabular}

%% file: generated/tab_qbank_confirm.tex
\begin{tabular}{@{}l l l c c c@{}}
\toprule
Layer & Set & Direction & 150 questions & 30 questions & Difference [95\%] \\
\midrule
20 & AISI held out & extended contrast & 0.974 & 0.971 & $+$0.0032 [$+$0.0003, $+$0.0064] \\
20 & AISI held out & selected detector & 0.963 & 0.948 & $+$0.0149 [$+$0.0111, $+$0.0192] \\
20 & Gender-Secret & extended contrast & 0.513 & 0.505 & $+$0.0086 [$-$0.0131, $+$0.0313] \\
20 & Gender-Secret & selected detector & 0.526 & 0.532 & $-$0.0052 [$-$0.0211, $+$0.0112] \\
36 & Gender-Secret & extended contrast & 0.444 & 0.409 & $+$0.0352 [$+$0.0018, $+$0.0684] \\
36 & Gender-Secret & selected detector & 0.751 & 0.761 & $-$0.0098 [$-$0.0225, $+$0.0027] \\
40 & AISI held out & extended contrast & 0.980 & 0.977 & $+$0.0029 [$+$0.0014, $+$0.0046] \\
40 & AISI held out & selected detector & 0.966 & 0.960 & $+$0.0062 [$+$0.0041, $+$0.0088] \\
40 & Gender-Secret & extended contrast & 0.491 & 0.467 & $+$0.0245 [$-$0.0045, $+$0.0559] \\
40 & Gender-Secret & selected detector & 0.778 & 0.802 & $-$0.0239 [$-$0.0373, $-$0.0098] \\
44 & AISI held out & extended contrast & 0.980 & 0.977 & $+$0.0024 [$+$0.0010, $+$0.0039] \\
44 & AISI held out & selected detector & 0.965 & 0.961 & $+$0.0042 [$+$0.0026, $+$0.0059] \\
44 & Gender-Secret & extended contrast & 0.605 & 0.541 & $+$0.0639 [$+$0.0409, $+$0.0897] \\
44 & Gender-Secret & selected detector & 0.808 & 0.807 & $+$0.0010 [$-$0.0105, $+$0.0126] \\
\bottomrule
\end{tabular}

%% file: refs.bib
@article{mahadik2026persona,
  title   = {Do Linear Probes Generalize Better in Persona Coordinates?},
  author  = {Mahadik, Prasad and Skapars, Adrians},
  journal = {arXiv preprint arXiv:2605.09391},
  year    = {2026}
}

@article{kretschmar2025liars,
  title   = {{Liars' Bench}: Evaluating Lie Detectors for Language Models},
  author  = {Kretschmar, Kieron and Laurito, Walter and Maiya, Sharan and Marks, Samuel},
  journal = {arXiv preprint arXiv:2511.16035},
  year    = {2025}
}

@inproceedings{goldowskydill2025detecting,
  title     = {Detecting Strategic Deception with Linear Probes},
  author    = {Goldowsky-Dill, Nicholas and Chughtai, Bilal and Heimersheim, Stefan and Hobbhahn, Marius},
  booktitle = {Proceedings of the 42nd International Conference on Machine Learning},
  series    = {Proceedings of Machine Learning Research},
  volume    = {267},
  pages     = {19755--19786},
  year      = {2025}
}

@article{zou2023representation,
  title   = {Representation Engineering: A Top-Down Approach to {AI} Transparency},
  author  = {Zou, Andy and Phan, Long and Chen, Sarah and Campbell, James and Guo, Phillip and Ren, Richard and others},
  journal = {arXiv preprint arXiv:2310.01405},
  year    = {2023}
}

@article{chen2025persona,
  title   = {Persona Vectors: Monitoring and Controlling Character Traits in Language Models},
  author  = {Chen, Runjin and Arditi, Andy and Sleight, Henry and Evans, Owain and Lindsey, Jack},
  journal = {arXiv preprint arXiv:2507.21509},
  year    = {2025}
}

@inproceedings{lu2026assistant,
  title     = {The {Assistant Axis}: Situating and Stabilizing the Default Persona of Language Models},
  author    = {Lu, Christina and Gallagher, Jack and Michala, Jonathan and Fish, Kyle and Lindsey, Jack},
  booktitle = {Proceedings of the 43rd International Conference on Machine Learning},
  series    = {Proceedings of Machine Learning Research},
  volume    = {306},
  pages     = {80955--81003},
  year      = {2026}
}

@inproceedings{kumar2026pressure,
  title     = {Pressure-Testing Deception Probes in {LLMs}: Scaling, Robustness, and the Geometry of Deceptive Representations},
  author    = {Kumar, Sachin},
  booktitle = {Proceedings of the Fifth Workshop on Generation, Evaluation and Metrics ({GEM})},
  pages     = {472--489},
  year      = {2026}
}

@inproceedings{kirch2025offpolicy,
  title     = {The Impact of Off-Policy Training Data on Probe Generalisation},
  author    = {Kirch, Nathalie and Dower, Samuel and Skapars, Adrians and Lubana, Ekdeep Singh and Krasheninnikov, Dmitrii},
  booktitle = {Findings of the Association for Computational Linguistics: {ACL} 2026},
  pages     = {22673--22729},
  year      = {2026}
}

@article{yoo2026subspace,
  title   = {Probe Generalization as Subspace Selection for {OOD} Deception Detection},
  author  = {Yoo, Daniel and Skapars, Adrians},
  journal = {arXiv preprint arXiv:2609.02893},
  year    = {2026}
}

@inproceedings{natarajan2026targeted,
  title     = {One Probe Won't Catch Them All: Towards Targeted Deception Detection},
  author    = {Natarajan, Vikram and Jain, Devina and Arora, Shivam and Golechha, Satvik and Bloom, Joseph Isaac},
  booktitle = {Proceedings of the 43rd International Conference on Machine Learning},
  series    = {Proceedings of Machine Learning Research},
  volume    = {306},
  pages     = {91932--91947},
  year      = {2026}
}

@article{ying2026truthfulness,
  title   = {The Truthfulness Spectrum Hypothesis},
  author  = {Ying, Zhuofan Josh and Ravfogel, Shauli and Kriegeskorte, Nikolaus and Hase, Peter},
  journal = {arXiv preprint arXiv:2602.20273},
  year    = {2026}
}

@inproceedings{azaria2023internal,
  title     = {The Internal State of an {LLM} Knows When It's Lying},
  author    = {Azaria, Amos and Mitchell, Tom},
  booktitle = {Findings of the Association for Computational Linguistics: EMNLP 2023},
  pages     = {967--976},
  year      = {2023}
}

@inproceedings{marks2023geometry,
  title     = {The Geometry of Truth: Emergent Linear Structure in Large Language Model Representations of True/False Datasets},
  author    = {Marks, Samuel and Tegmark, Max},
  booktitle = {Conference on Language Modeling (COLM)},
  year      = {2024}
}

@article{alain2016understanding,
  title   = {Understanding Intermediate Layers Using Linear Classifier Probes},
  author  = {Alain, Guillaume and Bengio, Yoshua},
  journal = {arXiv preprint arXiv:1610.01644},
  year    = {2016}
}

@inproceedings{pacchiardi2023catch,
  title     = {How to Catch an {AI} Liar: Lie Detection in Black-Box {LLMs} by Asking Unrelated Questions},
  author    = {Pacchiardi, Lorenzo and Chan, Alex J. and Mindermann, S{\"o}ren and Moscovitz, Ilan and Pan, Alexa Y. and Gal, Yarin and Evans, Owain and Brauner, Jan},
  booktitle = {International Conference on Learning Representations},
  year      = {2024}
}

@article{scheurer2023strategic,
  title   = {Large Language Models can Strategically Deceive their Users when Put Under Pressure},
  author  = {Scheurer, J{\'e}r{\'e}my and Balesni, Mikita and Hobbhahn, Marius},
  journal = {arXiv preprint arXiv:2311.07590},
  year    = {2023}
}

@inproceedings{li2024wmdp,
  title     = {The {WMDP} Benchmark: Measuring and Reducing Malicious Use with Unlearning},
  author    = {Li, Nathaniel and Pan, Alexander and Gopal, Anjali and Yue, Summer and Berrios, Daniel and Gatti, Alice and others},
  booktitle = {Proceedings of the 41st International Conference on Machine Learning},
  series    = {Proceedings of Machine Learning Research},
  volume    = {235},
  pages     = {28525--28550},
  year      = {2024}
}

@inproceedings{fiottokaufman2024nnsight,
  title     = {{NNsight} and {NDIF}: Democratizing Access to Open-Weight Foundation Model Internals},
  author    = {Fiotto-Kaufman, Jaden and Loftus, Alexander R. and Todd, Eric and Brinkmann, Jannik and Pal, Koyena and Troitskii, Dmitrii and others},
  booktitle = {International Conference on Learning Representations},
  year      = {2025}
}

@misc{taori2023alpaca,
  title        = {Stanford {Alpaca}: An Instruction-following {LLaMA} Model},
  author       = {Taori, Rohan and Gulrajani, Ishaan and Zhang, Tianyi and Dubois, Yann and Li, Xuechen and Guestrin, Carlos and Liang, Percy and Hashimoto, Tatsunori B.},
  howpublished = {\url{https://github.com/tatsu-lab/stanford_alpaca}},
  year         = {2023}
}

@misc{aisi_lie_rollouts,
  title        = {lie-detection-rollouts},
  author       = {{UK AI Security Institute}},
  howpublished = {Hugging Face dataset, \url{https://huggingface.co/datasets/ai-safety-institute/lie-detection-rollouts}, revision 177a816},
  year         = {2026},
  note         = {Accessed 22 September 2026}
}

@article{greenblatt2024alignment,
  title   = {Alignment faking in large language models},
  author  = {Greenblatt, Ryan and Denison, Carson and Wright, Benjamin and others},
  journal = {arXiv preprint arXiv:2412.14093},
  year    = {2024}
}

@article{levinstein2025still,
  title   = {Still No Lie Detector for Language Models: Probing Empirical and Conceptual Roadblocks},
  author  = {Levinstein, Benjamin A. and Herrmann, Daniel A.},
  journal = {Philosophical Studies},
  volume  = {182},
  number  = {7},
  pages   = {1539--1565},
  addendum= {arXiv:2307.00175},
  year    = {2025}
}

@inproceedings{buerger2024truth,
  title     = {Truth is Universal: Robust Detection of Lies in {LLMs}},
  author    = {B{\"u}rger, Lennart and Hamprecht, Fred A. and Nadler, Boaz},
  booktitle = {Advances in Neural Information Processing Systems},
  volume    = {37},
  addendum  = {arXiv:2407.12831},
  year      = {2024}
}

@article{moustafa2026beyond,
  title   = {Beyond {Liars' Bench}: The Impact of Lie Typology, Depth, and Sparsity on Deception Detection in {LLMs}},
  author  = {Moustafa, Amr and Feser, Max and Mai, Florian},
  journal = {arXiv preprint arXiv:2607.20479},
  year    = {2026}
}

@misc{marks2026persona,
  title        = {The Persona Selection Model: Why {AI} Assistants might Behave like Humans},
  author       = {Marks, Sam and Lindsey, Jack and Olah, Christopher},
  howpublished = {Anthropic Alignment Science Blog, \url{https://alignment.anthropic.com/2026/psm}},
  year         = {2026}
}

@inproceedings{joshi2024personas,
  title     = {Personas as a Way to Model Truthfulness in Language Models},
  author    = {Joshi, Nitish and Rando, Javier and Saparov, Abulhair and Kim, Najoung and He, He},
  booktitle = {Proceedings of the 2024 Conference on Empirical Methods in Natural Language Processing},
  pages     = {6346--6359},
  addendum  = {arXiv:2310.18168},
  year      = {2024}
}

@article{damour2022underspecification,
  title   = {Underspecification Presents Challenges for Credibility in Modern Machine Learning},
  author  = {D'Amour, Alexander and Heller, Katherine and Moldovan, Dan and others},
  journal = {Journal of Machine Learning Research},
  volume  = {23},
  number  = {226},
  pages   = {1--61},
  year    = {2022}
}

@article{geirhos2020shortcut,
  title   = {Shortcut Learning in Deep Neural Networks},
  author  = {Geirhos, Robert and Jacobsen, J{\"o}rn-Henrik and Michaelis, Claudio and Zemel, Richard and Brendel, Wieland and Bethge, Matthias and Wichmann, Felix A.},
  journal = {Nature Machine Intelligence},
  volume  = {2},
  number  = {11},
  pages   = {665--673},
  year    = {2020}
}

@article{park2023deception,
  title   = {{AI} Deception: A Survey of Examples, Risks, and Potential Solutions},
  author  = {Park, Peter S. and Goldstein, Simon and O'Gara, Aidan and Chen, Michael and Hendrycks, Dan},
  journal = {Patterns},
  volume  = {5},
  number  = {5},
  pages   = {100988},
  year    = {2024}
}

@article{jarviniemi2024uncovering,
  title   = {Uncovering Deceptive Tendencies in Language Models: A Simulated Company {AI} Assistant},
  author  = {J{\"a}rviniemi, Olli and Hubinger, Evan},
  journal = {arXiv preprint arXiv:2405.01576},
  year    = {2024}
}

@article{kadavath2022language,
  title   = {Language Models (Mostly) Know What They Know},
  author  = {Kadavath, Saurav and Conerly, Tom and Askell, Amanda and Henighan, Tom and Drain, Dawn and Perez, Ethan and Schiefer, Nicholas and Hatfield-Dodds, Zac and DasSarma, Nova and Tran-Johnson, Eli and others},
  journal = {arXiv preprint arXiv:2207.05221},
  year    = {2022}
}

@inproceedings{slobodkin2023curious,
  title     = {The Curious Case of Hallucinatory (Un)answerability: Finding Truths in the Hidden States of Over-Confident Large Language Models},
  author    = {Slobodkin, Aviv and Goldman, Omer and Caciularu, Avi and Dagan, Ido and Ravfogel, Shauli},
  booktitle = {Proceedings of the 2023 Conference on Empirical Methods in Natural Language Processing},
  pages     = {3607--3625},
  addendum  = {arXiv:2310.11877},
  year      = {2023}
}

@inproceedings{gottesman2024estimating,
  title     = {Estimating Knowledge in Large Language Models Without Generating a Single Token},
  author    = {Gottesman, Daniela and Geva, Mor},
  booktitle = {Proceedings of the 2024 Conference on Empirical Methods in Natural Language Processing},
  pages     = {3994--4019},
  addendum  = {arXiv:2406.12673},
  year      = {2024}
}

@inproceedings{ferrando2025know,
  title     = {Do {I} Know This Entity? {K}nowledge Awareness and Hallucinations in Language Models},
  author    = {Ferrando, Javier and Obeso, Oscar and Rajamanoharan, Senthooran and Nanda, Neel},
  booktitle = {International Conference on Learning Representations},
  addendum  = {arXiv:2411.14257},
  year      = {2025}
}

@article{cooney2026lie,
  title   = {``{Did} you lie?'' {Evaluating} Lie Detectors across Model Scale and Belief-Verified Model Organisms},
  author  = {Cooney, Alan and Africa, David and Irving, Geoffrey},
  journal = {arXiv preprint arXiv:2606.12618},
  year    = {2026}
}

@misc{aletheia2026quest,
  title        = {{Aletheia's Quest}: An {AI} Lie Detection Challenge},
  author       = {{Cadenza Labs} and {National Deep Inference Fabric} and {Schmidt Sciences}},
  howpublished = {\url{https://aletheias-quest.github.io/}},
  year         = {2026},
  note         = {Held-out results at \url{https://aletheias-quest.github.io/winners.html}}
}
